\documentclass[letterpaper, 10 pt, conference]{ieeeconf}  

\IEEEoverridecommandlockouts                              

\usepackage{graphics} 
\usepackage{epsfig} 
\usepackage{mathptmx} 
\usepackage{times} 
\usepackage{amsmath} 
\usepackage{amssymb}  
\usepackage[utf8]{inputenc}
\usepackage{url}
\usepackage{float}
\usepackage{subcaption}
\usepackage{cleveref}

\title{\LARGE \bf
An Open-source Hardware/Software Architecture and Supporting Simulation Environment to Perform Human FPV Flight Demonstrations for Unmanned Aerial Vehicle Autonomy
}

\author{Haosong Xiao$^{1}$, Prajit Krisshnakumar$^{2}$, Jagadeswara P K V Pothuri$^{1}$, \\ Puru Soni$^{3}$, Eric Butcher$^{3}$ and Souma Chowdhury$^{4}$
\thanks{$^\dagger$ Corresponding Author: soumacho@buffalo.edu}
\thanks{$^{1}$  Master's Student, Mechanical and Aerospace Engineering,
        University at Buffalo, Buffalo, NY 
        }%
\thanks{$^{2}$  PhD Student, Mechanical and Aerospace Engineering,
        University at Buffalo, Buffalo, NY 
        }%
\thanks{$^{3}$ Undergraduate Student, Computer Science and Engineering.,
University at Buffalo, Buffalo, NY }%
\thanks{$^{4}$ Associate Professor, Mechanical and Aerospace Engineering,
Co-Director, Center for Embodied Autonomy and Robotics,
University at Buffalo }%
\thanks{*This work was supported by Stephen Still Institute for Sustainable Transportation and Logistics (SSISTL) at University at Buffalo, the National Science Foundation (NSF) award CMMI 2048020, and the AFOSR DURIP award 13369473. Any opinions, findings, conclusions, or recommendations expressed in this paper are those of the authors and do not necessarily reflect the views of the SSISTL, NSF, and AFOSR.}
\thanks{*Copyright \textcopyright 2024 AIAA. Personal use of this material is permitted. Permission from AIAA must be obtained for all other uses, in any current or future media, including reprinting/republishing this material for advertising or promotional purposes, creating new collective works, for resale or redistribution to servers or lists, or reuse of any copyrighted component of this work in other works}
}

\begin{document}
\maketitle
\thispagestyle{empty}
\pagestyle{empty}

\begin{abstract}

Small multi-rotor unmanned aerial vehicles (UAVs), mainly quadcopters, are nowadays ubiquitous in research on aerial autonomy, including serving as scaled-down models for much larger aircraft such as vertical-take-off-and-lift vehicles for urban air mobility. Among the various research use cases, first-person-view (FPV) RC flight experiments allow for collecting data on how human pilots fly such aircraft, which could then be used to compare, contrast, validate, or train (via imitative learning) autonomous flight agents. While this could be uniquely beneficial, especially for studying UAV operation in contextually complex and safety-critical environments such as in human-UAV shared spaces, the lack of inexpensive and open-source hardware/software platforms that offer this capability along with low-level (complete access) to the underlying control software and data remains limited. To address this gap and significantly reduce barriers to human-guided autonomy research with UAVs, this paper presents an open-source software architecture implemented with an inexpensive in-house built quadcopter platform based on the F450 Quadcopter Frame. This setup uses two cameras to provide a dual-view FPV and an open-source flight controller, Pixhawk. The underlying software architecture, developed using the Python-based Kivy library, allows logging telemetry, GPS, control inputs, and camera frame data in a synchronized manner on the ground station computer. Since costs (time) and weather constraints typically limit numbers of physical outdoor flight experiments, this paper also presents a unique AirSim/Unreal Engine based simulation environment and graphical user interface (GUI) -- aka digital twin -- that provides a Hardware In The Loop (HITL) setup via the Pixhawk flight controller. We demonstrate the usability and reliability of the overall framework through a set of diverse physical FPV flight experiments and corresponding flight tests in the digital twin. 

\end{abstract}

\section{\bf INTRODUCTION}

Unmanned Aerial Vehicle (UAV) refers to a variety of flying equipment capable of performing tasks without an onboard human pilot. Initially introduced in the early 19th century for military use, quadcopters have evolved significantly with the advancement of sensors and microprocessors \cite{nonami2018research}. Modern UAVs range from small, electrically powered devices suitable for indoor navigation to larger aircraft capable of carrying passengers on long-range missions. Among UAVs, quadcopters are particularly popular due to their versatility and stability. 
The transportation sector greatly benefits from quadcopters, especially in Urban Air Mobility (UAM), last-mile delivery, and traffic monitoring, where their abilities in accurate GPS navigation, obstacle avoidance, and payload handling are indispensable. 
In the realm of entertainment, quadcopters elevate aerial filming with multi-UAV cooperation \cite{goh2021aerial}, introducing both redundancy and challenges like increased collision risks and the need for advanced obstacle avoidance \cite{pedersen2018neural}\cite{devos2018development}. Their role in localization, particularly in search and rescue missions \cite{ mcrae2019using}\cite{ behjat2021learning}, is pivotal, addressing crises in diverse terrains and demanding exceptional maneuverability \cite{mishra2020drone}. In construction, quadcopters are revolutionizing practices by enabling high-altitude visual inspections \cite{guo2023aerial}, pipeline monitoring \cite{marathe2019leveraging}, and site surveillance \cite{seo2018drone}. 
 While the adoption of quadcopters for these complex and dangerous operational tasks significantly reduces the likelihood of human injury, it concurrently elevates the demand for highly skilled First Person View (FPV) quadcopter pilots who can manage these complex tasks. Market research projects a substantial growth in this demand, anticipating a 51.1 percent increase from 2022 to 2027 \cite{VaughnCollege2022Drone}. In FPV flying, pilots control the quadcopters remotely from a first-person view using cameras mounted on them to navigate. In Urban Air Mobility (UAM) and last-mile delivery scenarios, FPV pilots expertly navigate through dense urban environments, where their ability to maneuver with precision is crucial for ensuring the safety of both people and the quadcopter themselves.
 
 The explosive advancement in artificial intelligence (AI) research further underscores this trend, shifting quadcopter operations towards increased autonomy and thereby diminishing the need for direct human pilot intervention \cite{floreano2015science}. This shift is particularly consequential for developing and operating electric Vertical Take-Off and Landing (VTOL) vehicles. AI's capability to process vast amounts of data in real-time potentially enhances the safety and reliability of these vehicles. For VTOLs, which are complex systems requiring precise control, AI algorithms can significantly improve navigation accuracy, energy efficiency, and overall operational effectiveness. This evolution marks a pivotal transition from traditional piloted systems to more autonomous, AI-driven solutions, reshaping the future of urban air mobility and beyond. Current research largely focuses on the design and control of VTOL systems \cite{ozdemir2014design, li2022dynamic, czyba2018construction}, operational efficiency, and safety measures, with few emerging studies examining the strategic use of reinforcement learning for air traffic control and scheduling within vertiports to optimize the deployment of eVTOLs in urban air mobility scenarios \cite{shao2021terminal, krisshnakumar2023fast, paul2024graph}.  
To effectively integrate autonomous capability into any aerial vehicle, including VTOL, its AI must be trained on extensive data derived from human pilots. This is critical for replicating human-like decision-making and providing a benchmark for autonomous flight systems. This ensures AI's adaptability and public trust \cite{knowles2021sanction, aoki2021importance}. 
Recent fields of research to align the performance of autonomous quadcopters with that of expert FPV quadcopter pilots refer to deep reinforcement learning \cite{li2017deep} and imitation learning \cite{hussein2017imitation}. The importance of deep reinforcement learning in quadcopter autonomy is crucial, particularly for autonomous decision-making in complex environments. This is evident in research by Wang et al. \cite{8993742}, Çetin et al. \cite{9081749}, and Anwar and Raychowdhury \cite{8978577}, which demonstrates the effectiveness of reinforcement learning in improving quadcopters' decision-making, especially in challenging settings like obstacle-rich areas or resource-limited environments. In the field of imitation learning, quadcopter learning from expert pilots significantly enhances maneuverability and efficiency.
Most of these methods used existing open-source flight data or expensive hardware to collect new data, limiting access to those with substantial resources. This barrier restricts smaller research teams and individual innovators who may not have the financial capability to invest in high-cost equipment. 
Further, current open-source datasets like EuRoC \cite{Burri25012016} and Zurich Urban MAV \cite{majdik2017zurich} provide robust image-based data with flight state information, but they are limited by their third-person view collection methodology. Moreover, frameworks used for autonomous FPV racing rely on indoor motion capture systems, and outdoor versions require external Leica trackers \cite{LeicaNovaMS60} for ground truth location, significantly posing challenges in terms of cost and spatial limitations.

Hence, the primary objective of this paper is to introduce an open-source framework equipped with FPV capabilities that is both cost-effective and readily accessible. This framework's software stack aims to enable precise capturing of human pilot data, facilitating the development of advanced AI systems. 
The platform is based entirely on the open-source flight controller - Pixhawk. Pixhawk supports multiple flying modes, accommodating FPV Quadcopter pilots of varying experience levels with its dual-view FPV user interface. In conjunction with this interface, Mavlink \cite{mavlink2023} communication will be established to fetch real-time quadcopter state and control signal data during FPV flights. Additionally, this data will be synchronously saved with dual-frame images and corresponding timestamps after each flight. The advantage of using Pixhawk as a flight controller is its flexibility in switching between quadcopter frames, allowing an extendable change from quadcopter FPV to fixed-wing FPV quadcopter at an acceptable cost compared to current market camera quadcopter models.
The secondary objective is to develop a digital twin platform (virtual environment) with Hardware-in-the-loop (HITL) capabilities and a GUI, providing an alternative solution for generating realistic, high-fidelity data, especially in scenarios where physical experimentation is expensive or challenging to perform very often \cite{kumar2021gui}. This digital twin approach allows for safe, repeatable, and controlled testing without the constraints associated with physical experiments.

The remainder of this paper is structured as follows: In Sec.~\ref{sec:framework}, we explain the components used in building the hardware, the software architecture and HITL simulation framework. Section \ref{sec:experiments} introduces the experiments performed to validate our framework, and in Sec.~\ref{sec:results}, we compare the results of experiments from both physical and digital twin approaches. Finally, in Sec.~\ref{sec:conclusion}, we provide our concluding remarks. 

\begin{figure}
    \centering
    \includegraphics[width=1\linewidth]{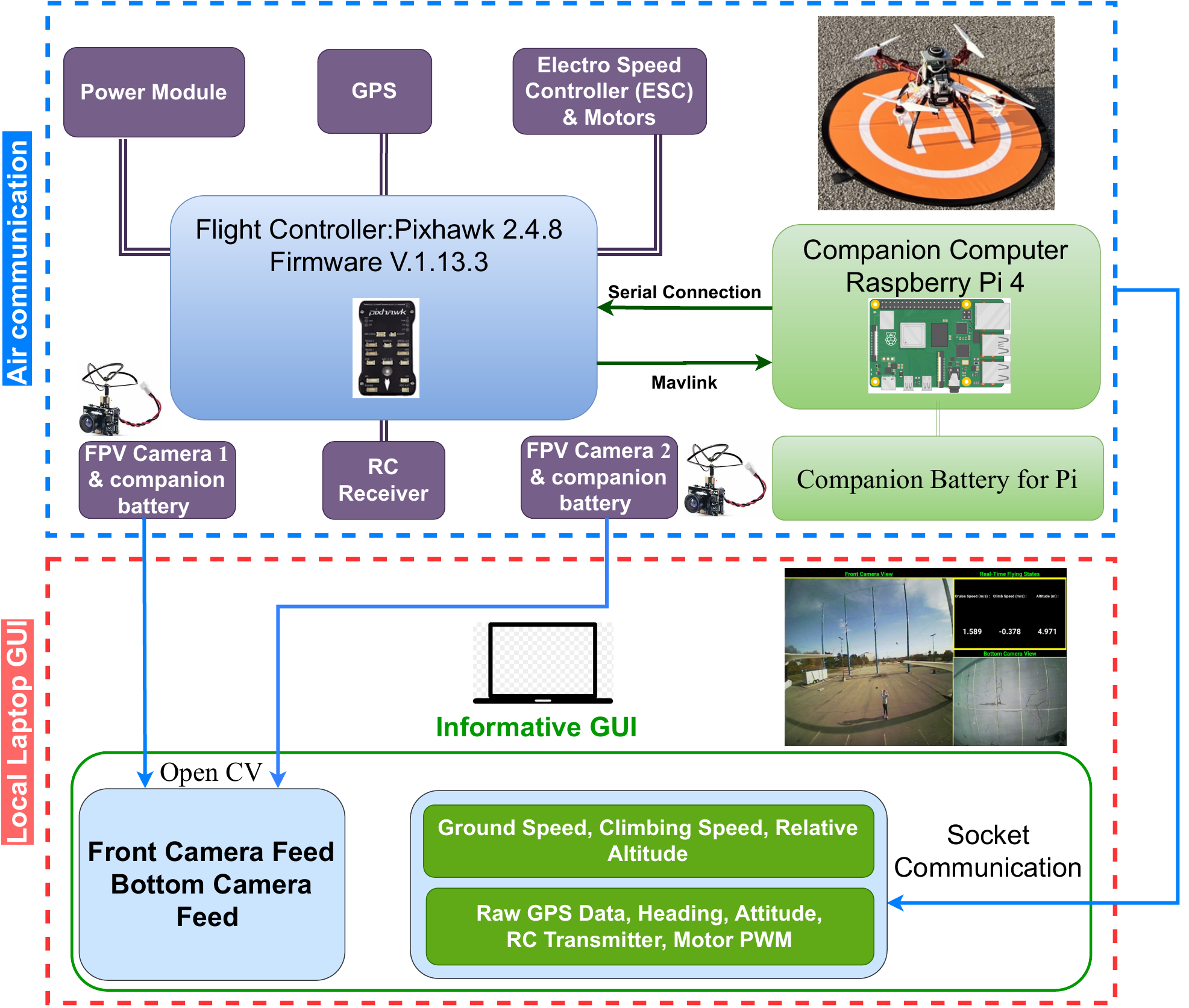}
    \caption{Overall Hardware Framework Illustrating the Connection and Flow of Data Between Various Modules}
    \label{fig:flowchart of framework}
\end{figure}

\begin{figure}
\includegraphics[width=1\linewidth]{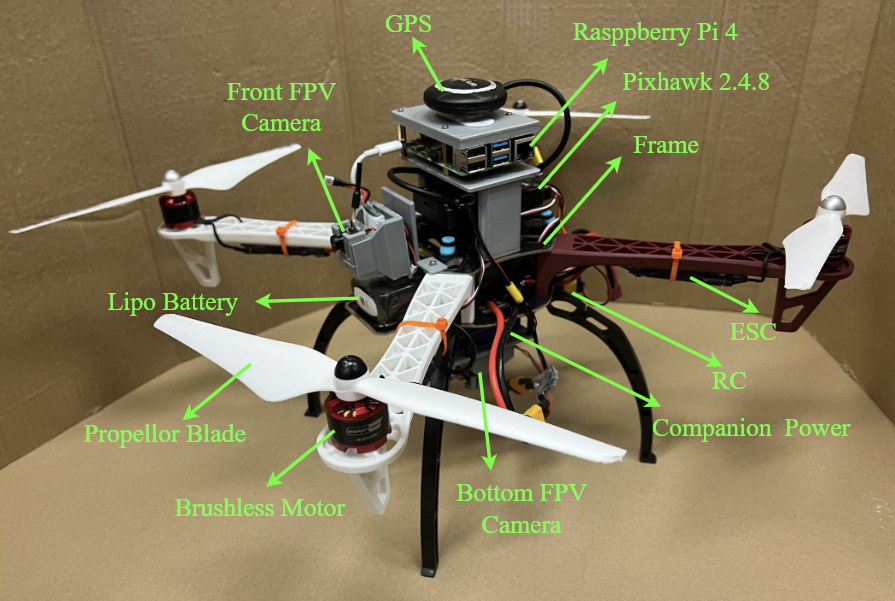}
\centering
\caption{In-house Built FPV Construction}
\label{fig:drone}
\end{figure}

\begin{table*}[h!]
\footnotesize
\centering
\caption{In-house Built FPV Component list}
\label{tab:hwcomponents}
\begin{tabular}{|c|c|c|c|c|cl}
\hline
\textbf{Component Name}    & \textbf{Count} & \textbf{Weight (g)} & \textbf{Dimensions (mm)}\\ \hline
 F450 Quadcopter Frame-kit                       & 1                & 280.00    &    360.00$\times$360.00$\times$180.00                   \\ \hline
 Readytosky 2212 920 Kv Brushless Motors    & 4                & 189.42     &     \(\varnothing 22.00 \times 12.00\)
                 \\ \hline
 Hobbypower Simonk 30A ESC                  & 4                & 103.76     &      50.00$\times$25.00$\times$10.00                 \\ \hline
 Dilwe 9.4 inch Self-Lock Propelloer Blades & 4                & 45.36      &    $ \varnothing 238.76$ \                 \\ \hline
 Pixhawk 2.4.8 Flight Controller Pack       & 1                & 40.00      &    80.00$\times$50.00$\times$16.00                 \\ \hline
 M8N GPS                                    & 1                & 54.43      &    58.00$\times$58.00$\times$12.00                   \\ \hline
 Raspberry Pi 4 (8GB Ram Version)           & 1                & 143.17     &     85.60$\times$56.50$\times$29.90               \\ \hline
 Wolfwhoop 600VTL FPV Camera                & 2                & 10.00      &     20.00$\times$14.00$\times$34.00                   \\ \hline
 Ovonic 5500 mAh 3S1P Lipo Battery          & 1                & 376.00     &    154.00$\times$46.00$\times$23.00                   \\ \hline
 SunFounder Raspberry Pi Power (5V/3A)      & 1                & 170.10     &    85.60$\times$56.50$\times$50.00                   \\ \hline
 URGENEX 3.7V 1S Lipo Battery               & 2                & 1.71       &      43.80$\times$24.60$\times$9.60                 \\ \hline
 Frsky i6X RC Receiver and Transmitter      & 1                & 16.30 (receiver)        & 46.99$\times$26.16$\times$16.99         \\ \hline
\end{tabular}
\end{table*}

\begin{table}[]
\footnotesize
\centering
\caption{In-house Built FPV Performance Parameter}
\label{tab: hwpp}
\begin{tabular}{|c|c|c|}
\hline
\textbf{Mass (g)} & \textbf{Hover Endurance (mins)} & \textbf{Suggested Flying Speed (m/s)}  \\ \hline
$\sim$ 1522            &  $\sim$ 11                        & 1.20                                      \\ \hline
\end{tabular}
\end{table}

\section{\bf Hardware and Simulation Framework}\label{sec:framework}
The in-house built FPV quadcopter framework, as shown in figure~\ref{fig:flowchart of framework}, requires a finely-tuned FPV platform and an informative Graphical User Interface (GUI) to support pilots by providing real-time FPV camera views and flying state data such as ground speed, climbing speed, and relative altitude. We also implemented Airsim HITL with GUI in Unreal Engine to support FPV flight in virtual environment. The below subsections will deep into the development of hardware and software platforms. For simplicity, in later sections, the in-house built FPV quadcopter will be referred to as the physical quadcopter, and the Airsim HITL platform with compiled GUI will be referred to as the digital twin.

\subsection{\bf Hardware Platform }
This section outlines physical quadcopter construction. The final product is shown in figure \ref{fig:drone}, and it includes a frame, brushless motors, propeller blades, battery, electronic speed controller (ESC), RC receiver, transmitter, flight controller, companion computer, and two cameras. Table \ref{tab:hwcomponents} lists the hardware components used in the platform as well as the weights of each component, while table \ref{tab: hwpp} lists the performance parameters.
\subsubsection{Frame}
The frame of the physical quadcopter utilizes the F450 model from YoungRC, selected for its high-strength arms, which provide a balance of durability and weight management. The frame includes an integrated PCB board, streamlining the soldering process for ESCs and power modules and thereby improving the reliability of connections. Furthermore, the frame's large mounting tab on the bottom plate affords ample space for mounting a bottom-facing camera, enhancing support for FPV flight and precision in landing tasks. 
\subsubsection{Motors and Propellers}
Considering the chosen frame for the physical quadcopter, ESCs, motors, and propellers were carefully selected to ensure sufficient thrust, smooth flight performance, and efficient power consumption without overloading the motors. A set of 920 Kv brushless motors paired with 9.4-inch propeller blades was chosen, taking into account the frame's size and the quadcopter's agility requirements.
\subsubsection{Battery and ESCs}
Based on the voltage regulation specified for the selected motors, the recommended operating voltage range is 7V to 12V. Consequently, a 3-cell LiPo battery, with each cell providing 3.7V, has been selected as the power source to support the quadcopter's flight, yielding a total nominal voltage of 11.1V. To provide sufficient redundancy and prevent motor overheating, 30A Electronic Speed Controllers (ESCs) have been chosen for implementation in the FPV quadcopter.
\subsubsection{Flight controller and companion computer}
The flight controller is a critical component within this system, particularly given the aim of this study to develop a methodology for bridging the gap between the physical quadcopter framework and the digital twin. The Unreal Engine has achieved high-fidelity quadcopter simulation using AirSim's HITL capability. Therefore, a flight controller that supports both HITL configurations and actual flight operations is essential. The Pixhawk 2.4.8 has been selected as the flight controller for the FPV quadcopter due to its affordability and capability to provide smooth flight control. Given the limitations of telemetry in the transmission rate of flight data, a Raspberry Pi has been integrated as a companion computer with the Pixhawk via a serial connection. This setup facilitates further development of the FPV GUI and enhances data collection capabilities.
\subsubsection{FPV Cameras, RC Transmitter and Receiver}
To enable a dual-view FPV flying experience, two FPV cameras with 600 TV lines (TVL) were selected as image sources. A crucial factor in integrating FPV cameras with an RC transmitter and receiver, especially considering the range expectation of 200 meters minimum, is the support range. Wolfwhoop 600 TVL FPV cameras and the FrSky i6X RC transmitter and receiver have been chosen for this FPV physical platform to ensure reliable control quality and simultaneous image feeds during flight within this range. Please note that to successfully receive images from the FPV camera transmitter, a compatible radio receiver is also needed to connect to a local laptop.
\subsection{\bf Graphical User Interface}
\begin{figure} [H]
\includegraphics[width=1\linewidth]{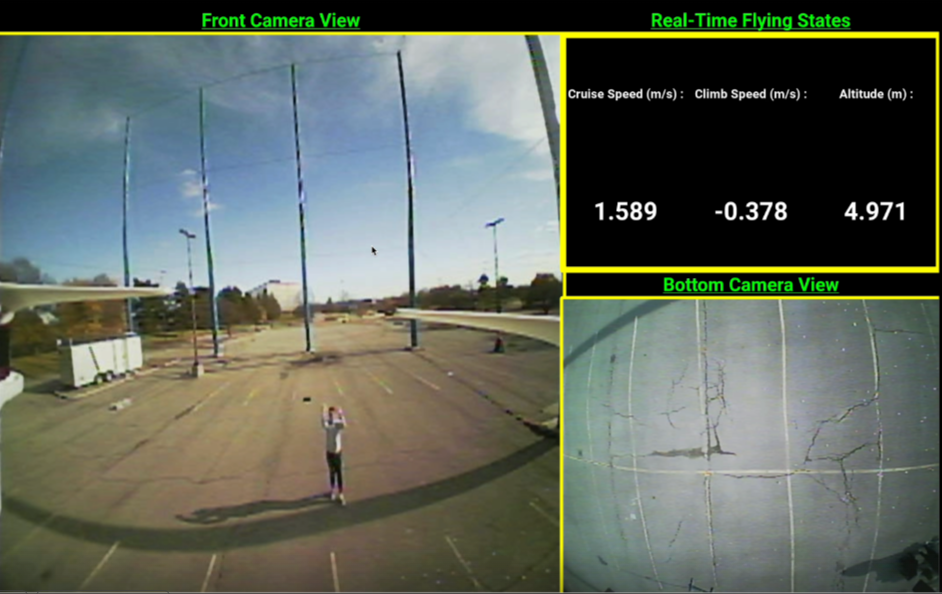}
\centering
\caption{Dual-View GUI Screen Shot}
\label{fig:GUI}
\end{figure}
To facilitate smooth FPV (First-Person View) flight, an informative GUI supporting dual-view FPV flying has been developed using the Kivy library \cite{kivy}, a robust open-source Python framework designed for cross-platform GUI development. Reflecting on the construction of the physical quadcopter, the dual cameras and the flight controller are identified as the primary sources of information. These are concurrently displayed on the GUI during flight, providing critical real-time data.
The dual-view GUI framework is adeptly designed to manage functionalities such as real-time flying state acquisition via local network communication and processing of image frames from both front and bottom cameras. The interface's primary division is a box layout in horizon distribution: On the left, a larger image frame is for the front camera view display, and on the right is a vertical stack of two box widgets containing three real-time flying states and the bottom camera view. 
To address the discrepancy in refresh rates between the sensors and FPV cameras, the GUI framework leverages Kivy's non-blocking clock events updates. This allows each data source to update with their unique fresh rate in the same thread. It also employs multiprocessing to prevent the main thread of the GUI app from being blocked in the sensor data reading process. Additionally, considering the limitations in data refresh rates through telemetry communication, the GUI framework integrated socket communication to receive data sent from Raspberry Pi, which receives Pixhawk data through the serial connection. In the air, a Raspberry Pi connects to the Pixhawk flight controller through GPIO pins, runs a Python script to read the real-time flight data via Pymavlink \cite{Pymavlink2023}, and then sends it to the local GUI as the server using the Python script. Meanwhile, the GUI on the local machine runs as the client, receiving the data sent from the server on the Raspberry Pi. This data transmission process is handled separately through multiprocessing, allowing it to be retrieved recursively through a callback function without blocking the main thread. The codes for software stacks are available in our github.\footnote{\url{https://github.com/adamslab-ub/FPV_quadcopter_aviation.git}}

\subsubsection{GUI Main Structure}
\label{sec:class}
The core structure of the GUI framework comprises two independent processes: one for data fetching from the quadcopter via local network communication with the onboard computer, Raspberry Pi, and the other for the main application process. This process constructs the overall GUI layout and integrates all functionalities into a single interface. All functionalities are programmed using object-oriented programming (OOP), ensuring a structured and efficient approach to managing the complex interactions within the GUI. The main application process is carried out through three classes, with specific tasks: \\
\begin{itemize}
    \item \lq{Datatable}\rq{} Class: This class fetches, displays, and logs data in the GUI. It starts an independent process for socket communication, pushing real-time Pixhawk data sent from the Raspberry Pi to the main thread. Once the communication process is established, the callback function within the class is recursively called via Kivy's Clock object to update and log the data transmitted from the other process per iteration, along with the corresponding image frames. Additionally, speed and relative altitude information, retrieved and computed from the updated data, are displayed on the main GUI frame's widget in a grid layout.
    \item \lq{Camview}\rq{} Class: The \lq{Camview}\rq{} class is responsible for video streaming. It sets up the procedures from image capture to stream on the widget with desired FPS, through OpenCV \cite{opencv_library} and schedules frame updates via Kivy’s Clock object. The class captures, resizes, and processes image frames into textures for display on the widget, retrieves the latest captured frame for synchronous logging, and releases camera resources when the app shuts down. This ensures proper management and display of real-time video feeds without lagging.
    \item \lq{MainGUI}\rq{} Class: As the main application class, it arranges the overall GUI layout and assigns the widgets for functionalities. It initiates functionalities by calling the \lq{Camview}\rq{} class twice to set up the front and bottom streaming, along with the \lq{Datatable}\rq{} class for real-time data fetching. The main class adds widgets for each of the initiated functionality instances, placing these widgets in the corresponding horizontal box layout. Besides, the main GUI class initializes the folder path directory, using the timestamp as the name of the path, and passes through the \lq{Datatable}\rq{} class for further data logging manipulation. This setup demonstrates the composition nature of OOP, where the \lq{MainGUI}\rq{} class manages and coordinates the functionality of other classes.
\end{itemize}

\subsubsection{Flying State Data Fetching and Data Collection}
 Recalling the chosen flight controller is the Pixhawk 2.4.8, which communicates using the Mavlink \cite{mavlink2023} protocol. Two methods were considered for building the data transmission bridge: telemetry communication and a serial connection. The serial connection with a companion computer was selected due to its superior data transfer rate, ensures a robust and direct link between the Pixhawk and the companion computer, resulting in more efficient data handling.
 In this underlying GUI software setup, the data being transferred and saved includes raw GPS data, speed data, RC control PWM data, servo PWM data, and attitude data.  One noticeable detail regarding altitude shown in the table widget is that the altitude displayed is the result of the real-time altitude minus the first altitude recorded when the quadcopter is armed, in units of meters. Compared to directly negating the relative altitude read through Pymavlink \cite{Pymavlink2023}, this method helps ignore error readings caused by the initial powering position. \\
 
 The synchronous data collection is facilitated through a collaborative effort between the functionality classes introduced in section \ref{sec:class}. The \lq{MainGUI}\rq{} class initializes the data file and image folder directories and passes them to the \lq{Datatable}\rq{} class. Once real-time data from the Pixhawk starts being read by the Raspberry Pi, it is sent through socket communication from the Pi end as the server. The function in the GUI responsible for socket communication, running as the client, is passed to the multiprocessing object. In \lq{Datatable}\rq{} class, once the multiprocessing starts after being initialized, the bridge for data transmission between the Raspberry Pi and the GUI is officially built and ready to be queued by the callback function at each clock schedule interval. Subsequently, the callback function retrieves the current timestamp and logs the received data, along with image frames obtained by invoking the \lq{getframe}\rq{} method of the \lq{Camview}\rq{} class, into predefined directories for each callback iteration until the flight is complete.

\subsection{\bf AirSim Simulation Environment} 
\subsubsection{Introduction to Simulation Environment}
Microsoft AirSim \cite{shah2018airsim} is an open-source robotics simulation platform. AirSim helps us solve the need for large data sets for training and allows debugging in a simulator. AirSim leverages current game engine rendering, physics, and perception computation to create accurate, real-world simulations. Together, this realism, based on efficiently generated ground-truth data, enables the study and execution of complex, time-consuming, and risky missions in the real world.
The default quadcopter model in AirSim is modified to match our real hardware; the morphology and the cameras are modified to match the physical quadcopter. Airsim allows HITL simulation with PX4, an open-source, popular flight controller. HITL simulation offers a blend of simulation safety and real-world unpredictability. While software simulations are insightful, HITL simulation captures the nuances of actual hardware. 
At the same time, the HITL simulation starts, a python file will run to help retrieve the digital twin's flying states along with image frames from two views, front and bottom, in corresponding timestamps simultaneously through Mavlink \cite{mavlink2023} and Airsim API \cite{shah2018airsim}.
\begin{figure} 
    \centering
    \includegraphics[width=1\linewidth]{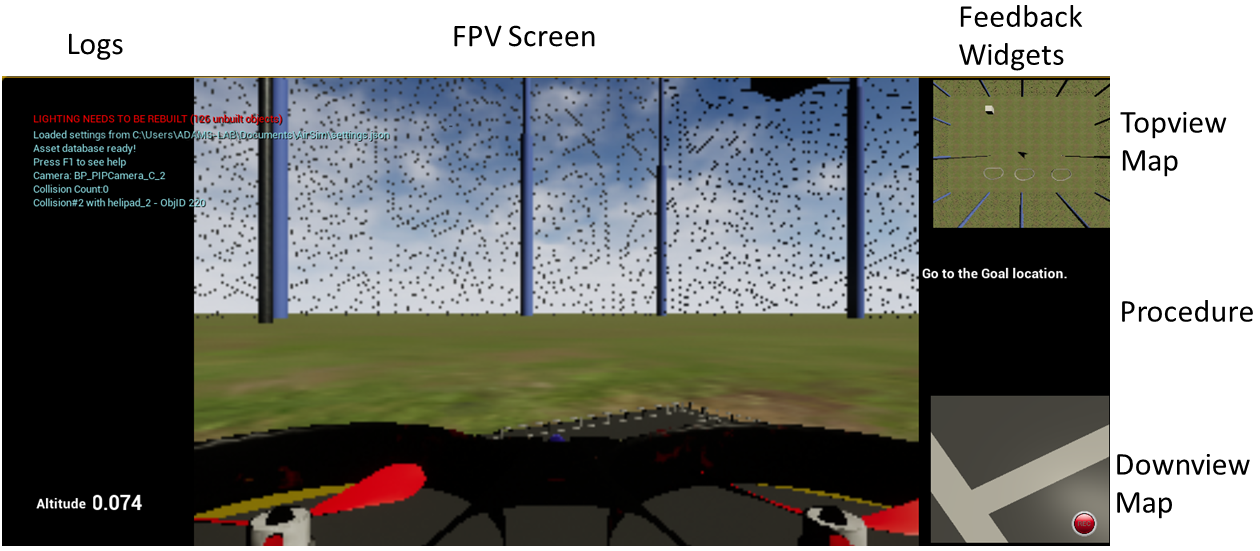}
    \caption{Graphical User Interface for FPV simulation based on Unreal Engine}
    \label{fig:sim_gui}
\end{figure}
A virtual version of the 24,000-square-foot Structure of Outdoor Autonomy Research (SOAR) facility at the University at Buffalo has been implemented through Unreal Engine 4.27. The environment is built precisely to match the real-world dimensions. In this virtual environment, pilots can free roam around using a graphic user interface (GUI) which includes two camera feeds, front view and bottom view, altitude information, and a top view global map, or there are few pre-defined tasks that the researchers can use while collecting data. Unreal Engine allows easy drag-and-drop features for creating a new task, which helps generate various tasks quickly to generate data. 
\begin{figure} [H]
    \centering
    \includegraphics[width=1\linewidth]{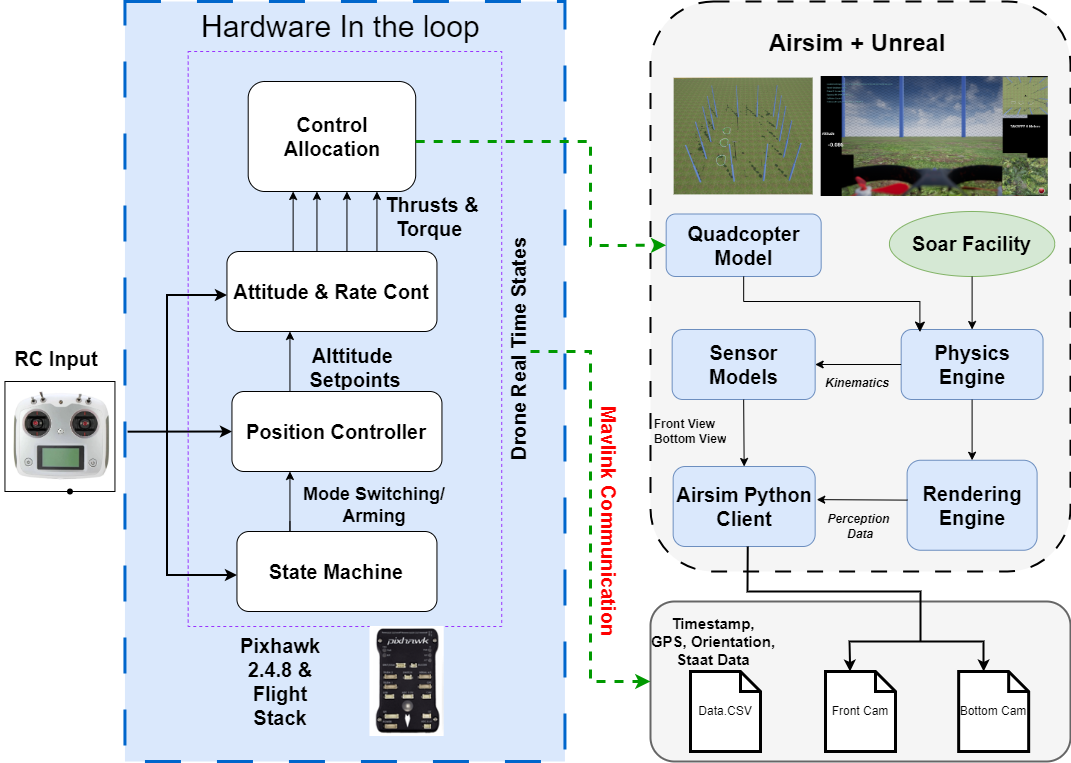}
    \caption{Flowchart for FPV Simulation based on Unreal Engine}
    \label{fig:simulation flowchartt}
\end{figure}

The same hardware controller, Pixhawk, will be used for the digital twin's controller in the virtual environment, while it's not controlling a real quadcopter kit set-up, it is interfaced with a virtual quadcopter with virtual sensors simulated from Airsim. For user ends, the pilot will control the digital twin in a simulation environment through embedded GUI shown in Fig.~\ref{fig:sim_gui}; third-person view flying is also available to switch. The simulation GUI consists of more feedback panels for the predefined tasks than the physical quadcopter platform. 3 widgets are added to the right side of the screen to show the users a top-view map to know their relative location, a procedure that shows their current task, and a downview camera similar to the physical quadcopter platform. 
Figure \ref{fig:simulation flowchartt} shows the flowchart of simulation environment. 

\subsubsection{Flying State Data Collection}

The Pixhawk processes virtual sensor data encompassing various quadcopter states, including raw GPS data, orientation, cruise speed, and climb rate. The pilot can observe this data as it manifests in the digital twin's flying position and attitude. The pilot interacts by maneuvering the RC joystick, which is integrated with the RC receiver connected to the Pixhawk. The data collection for the digital twin requires collaboration between Mavlink communication and the Airsim API. This collaboration captures flying state data, including simulated raw GPS, attitude, speed, and joystick signal input, which is then retrieved via Mavlink communication and saved in a CSV file. Simultaneously, corresponding front and bottom images captured via the AirSim API are saved as PNG files in separate folders within the same directory.

\section{\bf Physical Experiment and Demonstrations}\label{sec:experiments}
\begin{figure} [H]
    \centering
    \begin{subfigure}[b]{0.48\columnwidth}
        \centering
        \includegraphics[width=\textwidth]{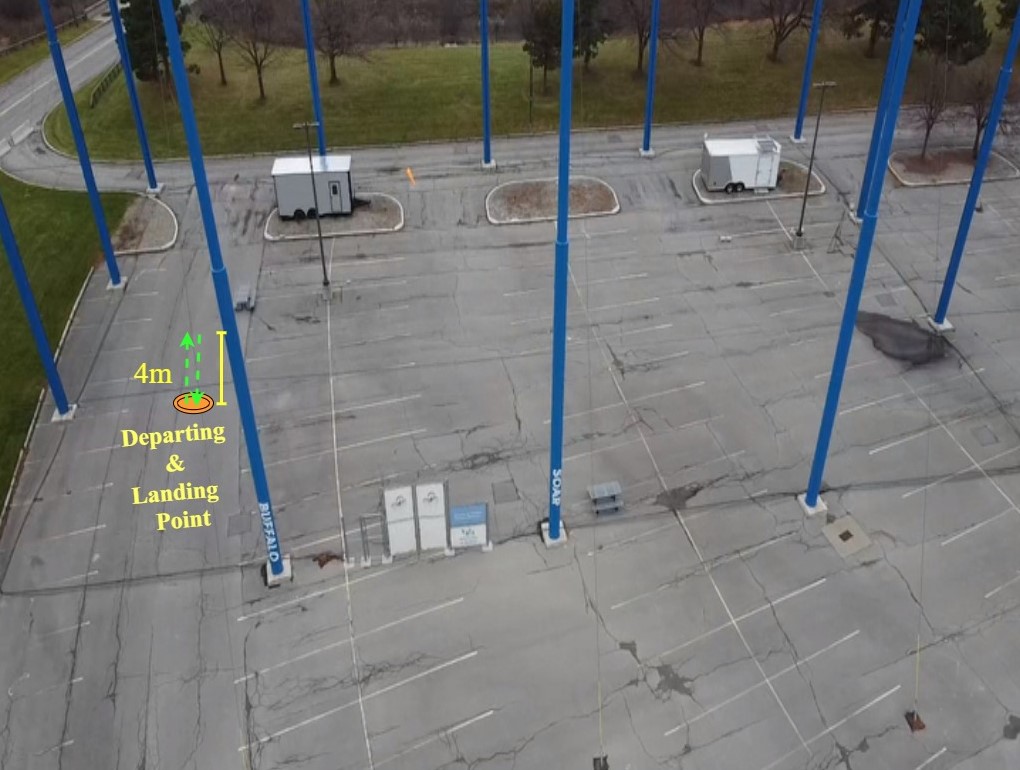}
        \caption{Task 1: Take off, hover, and land}
        \label{fig:task1}
    \end{subfigure}
    \hfill
    \begin{subfigure}[b]{0.48\columnwidth}
        \centering
        \includegraphics[width=\textwidth]{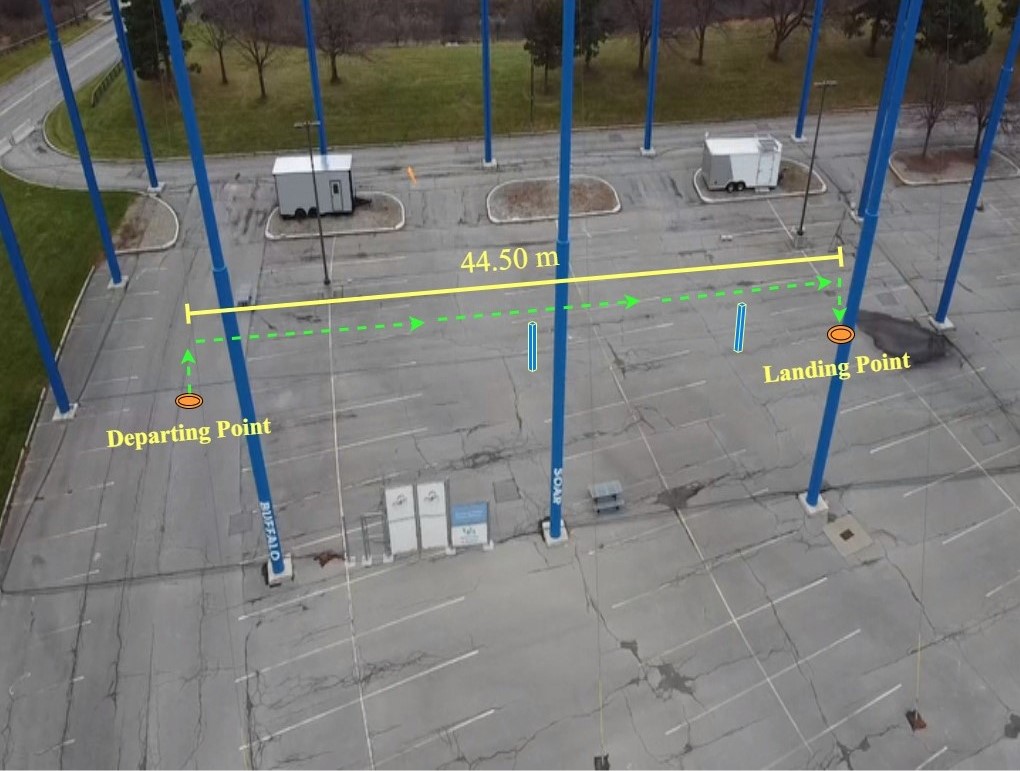}
        \caption{Task 2: Flying from point A to B}
        \label{fig:task2}
    \end{subfigure}
    \caption{Flying environment at SOAR for task 1 and task 2}
    \label{fig:flying task 1-2}
\end{figure}

\begin{figure} [H]
    \centering
    \begin{subfigure}[b]{0.48\columnwidth}
        \centering
        \includegraphics[width=\textwidth]{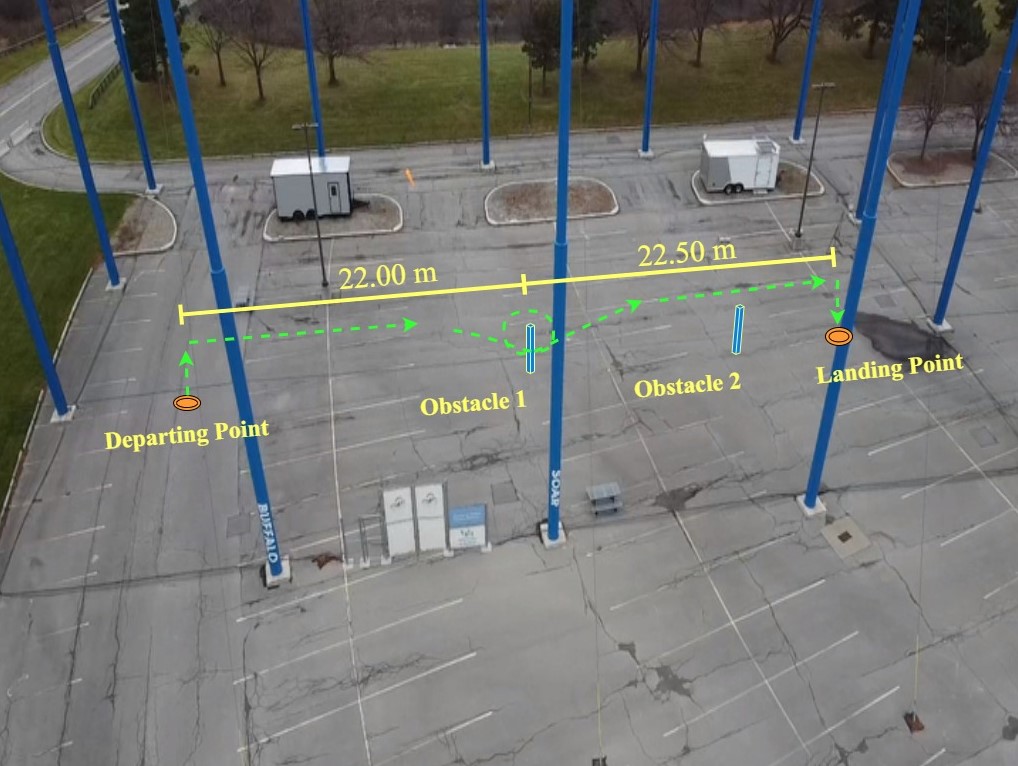}
        \caption{Task 3: Obstacle avoidance}
        \label{fig:task3}
    \end{subfigure}
    \hfill
    \begin{subfigure}[b]{0.48\columnwidth}
        \centering
        \includegraphics[width=\columnwidth]{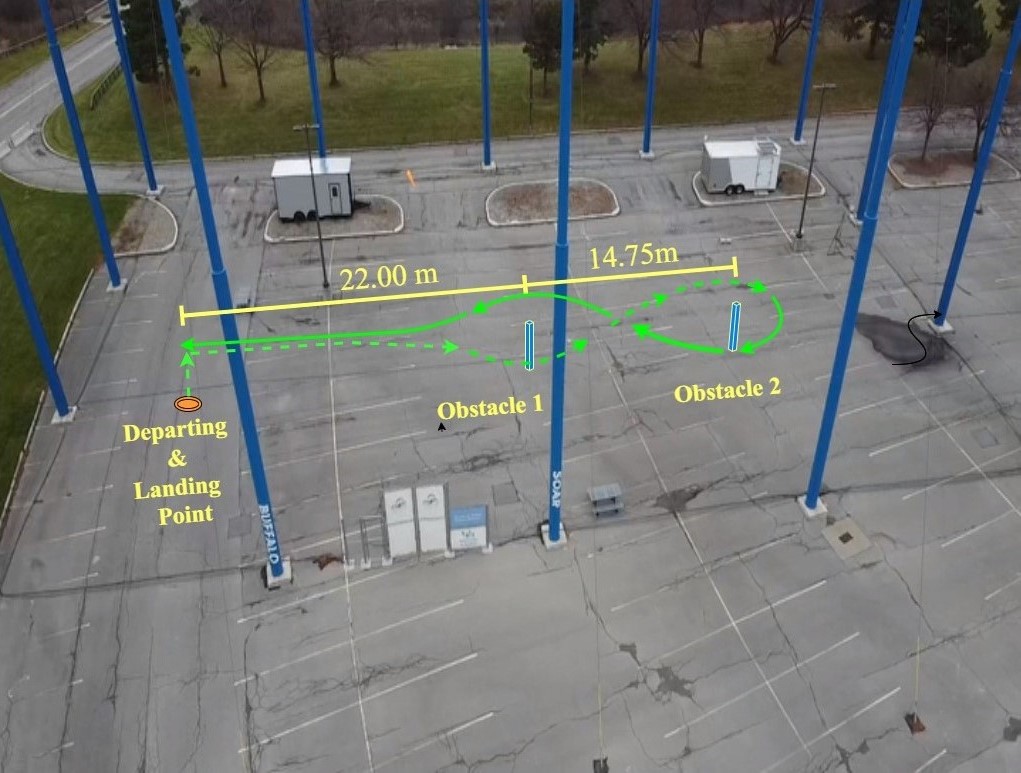}
        \caption{Task 4: Flying Figure 8}
        \label{fig:task4}
    \end{subfigure}
    \caption{Flying environment at SOAR for task 3 and task 4}
    \label{fig:flying task 3-4}
\end{figure}

This section outlines physical experiments to validate the physical quadcopter platform at the Structure of Outdoor Autonomy Research (SOAR) at the University at Buffalo. Throughout these experiments, one single pilot utilized the physical quadcopter to complete four flying tasks, as indicated in subsection A, Figure \ref{fig:flying task 1-2}--\ref{fig:flying task 3-4}. The flying environment involved two 3-meter obstacles aligned in line between the starting and landing spots. This distribution was then replicated in a simulation environment. The same pilot also employed the virtual GUI in this simulation environment, using the FPV view, to complete identical tasks. This process aimed to collect flying state data through Airsim HITL, thus validating the dataset collection through the digital twin platform.  Results will be collected from both platforms using the same 'Position' flight mode \cite{PX4position} for a fair comparison. The complexity of the flying tasks was ascent so that the pilot's trajectory for higher complexity tasks is expected to show more flying position adjustment.

\subsection{\bf Experiment tasks} \label{sec: tasks}
\subsubsection{Flying Task 1: Take off, hover and land}
In Task 1, the pilot was required to perform a series of maneuvers, including taking off vertically and ascending to a height of 4 meters. Upon reaching this altitude, the task entailed hovering above the helipad. This required the pilot to adjust the pitch and roll of the quadcopter based on the bottom image feed. The quadcopter was expected to maintain this hover for 10 seconds before landing back at the original departure spot.
\subsubsection{Flying Task 2: Flying from point A to point B} 
Similar to the beginning of Task 1, for Task 2, the pilot was asked to perform a series of maneuvers that included taking off vertically and ascending to an altitude above 3 meters. Once reaching an altitude of 3 meters, the pilot then maneuvered the quadcopter above the obstacles straight to the landing spot. As soon as the landing marker appeared in the bottom view, the pilot was required to halt and execute a slow descent onto the landing spot.


\subsubsection{Flying Task 3: Obstacle avoidance} 
Flying Task 3 required the pilot to navigate from the departure spot to the landing spot, similar to Task 2, but with the added challenge of obstacle avoidance. This task began with the pilot taking off and ascending to an altitude of 3 meters. However, unlike the previous task, upon reaching the midpoint of their route, pilots were tasked with an additional maneuver: circling the first obstacle. After completing the circular navigation around the first obstacle, the pilot maneuvered above the second obstacle and continued towards the landing spot, where he executed a slow descent to land. 
\subsubsection{Flying task 4: Flying Figure 8}
Flying task 4 is the most difficult task considering the number of maneuvering requirements. The pilot started by taking off and ascending to an altitude of 3 meters, cruised to the first obstacle, and circled the two obstacles in the shape of Figure 8. Once the circulation maneuver was finished, the pilot then maneuvered back to the departing spot and slowly descended to land.

\subsection{\bf Geo-coordinate conversion}
\label{sec:GPScoordinate}
The process of path visualization was carried out through the conversion of geographic coordinates. Given the raw GPS data collected from the dataset via the physical quadcopter GUI platform, this data was initially converted to Earth-Centered Earth-Fixed (ECEF) coordinates. This conversion generates the global position of each recorded position in the flying trajectory relative to the Earth's center of mass. Using the initial starting point from the trajectory points, post-ECEF transformation, as the reference, all ECEF coordinates are then transformed into the East-North-Up (ENU) coordinate system. This transformation is essential for visualizing the trajectory path of the flights. To better visualize the flown trajectory comparison between the physical quadcopter and the digital twin, a reference coordinate aligned with the starting point from each environment was selected. The line drawn between the starting point and the reference point was then ensured to be straight and aligned with the pure east axis in the ENU frame.

\section{\bf Results and Discussions}\label{sec:results}

\begin{figure} [H]
    \centering
    \begin{subfigure}[b]{0.48\columnwidth}
        \centering
        \includegraphics[width=\columnwidth]{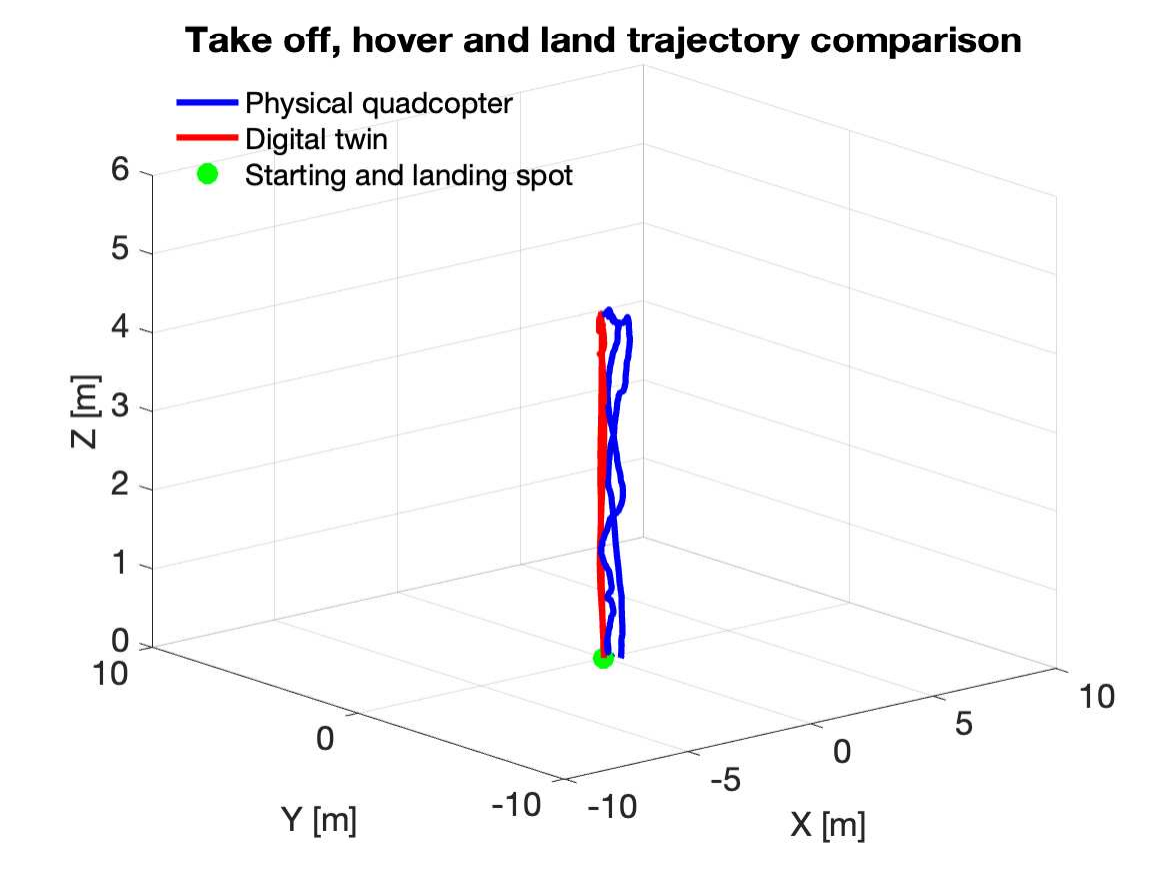}
        \caption{3D trajectory of Task 1 (both platforms)}
        \label{fig:3Dtask1}
    \end{subfigure}
    \hfill
    \begin{subfigure}[b]{0.48\columnwidth}
        \centering
        \includegraphics[width=\columnwidth]{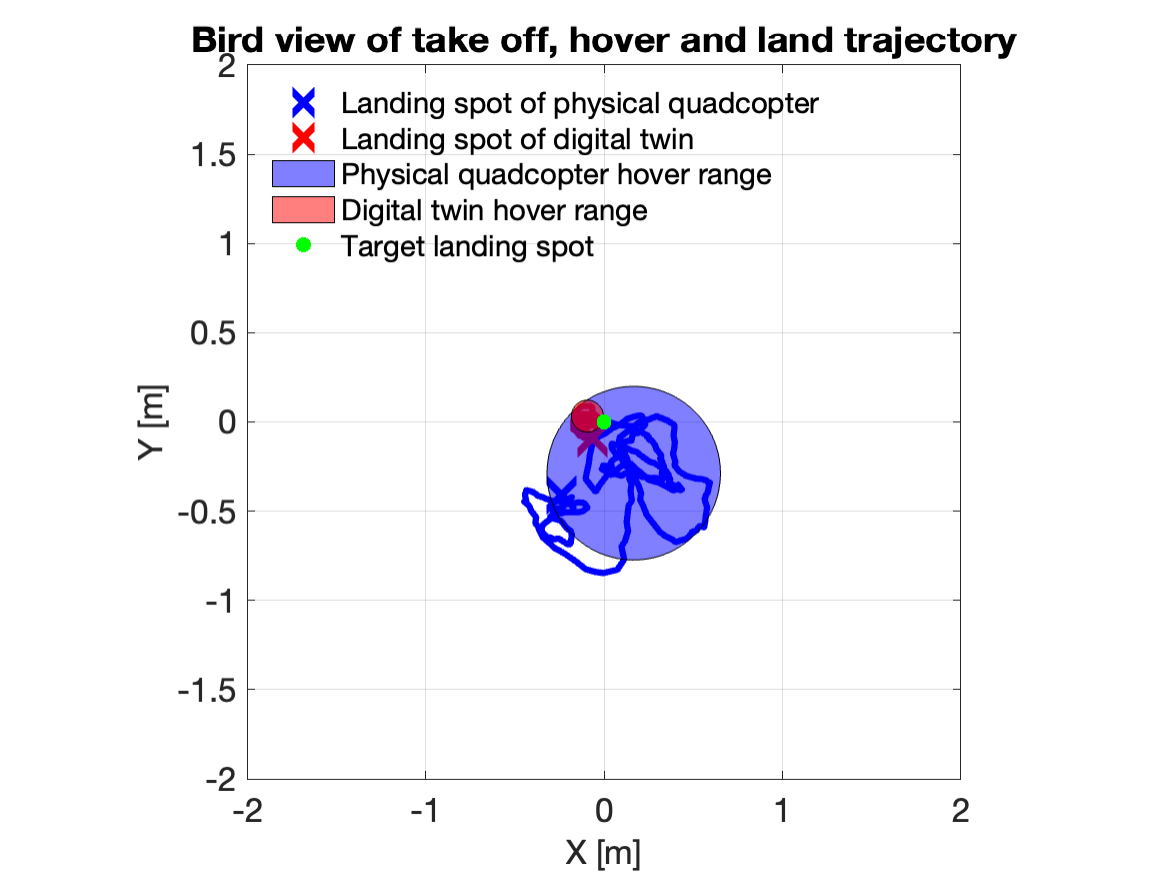}
        \caption{Bird-view trajectory of Task 1 (both platforms)}
        \label{fig:2Dtask1}
    \end{subfigure}
    \caption{Visualized flown trajectory of Task 1 using both platforms}
    \label{fig:task1result1/2}
\end{figure}

\begin{figure} [H]
    \centering
    \begin{subfigure}[b]{0.48\columnwidth}
        \centering
        \includegraphics[width=\columnwidth]{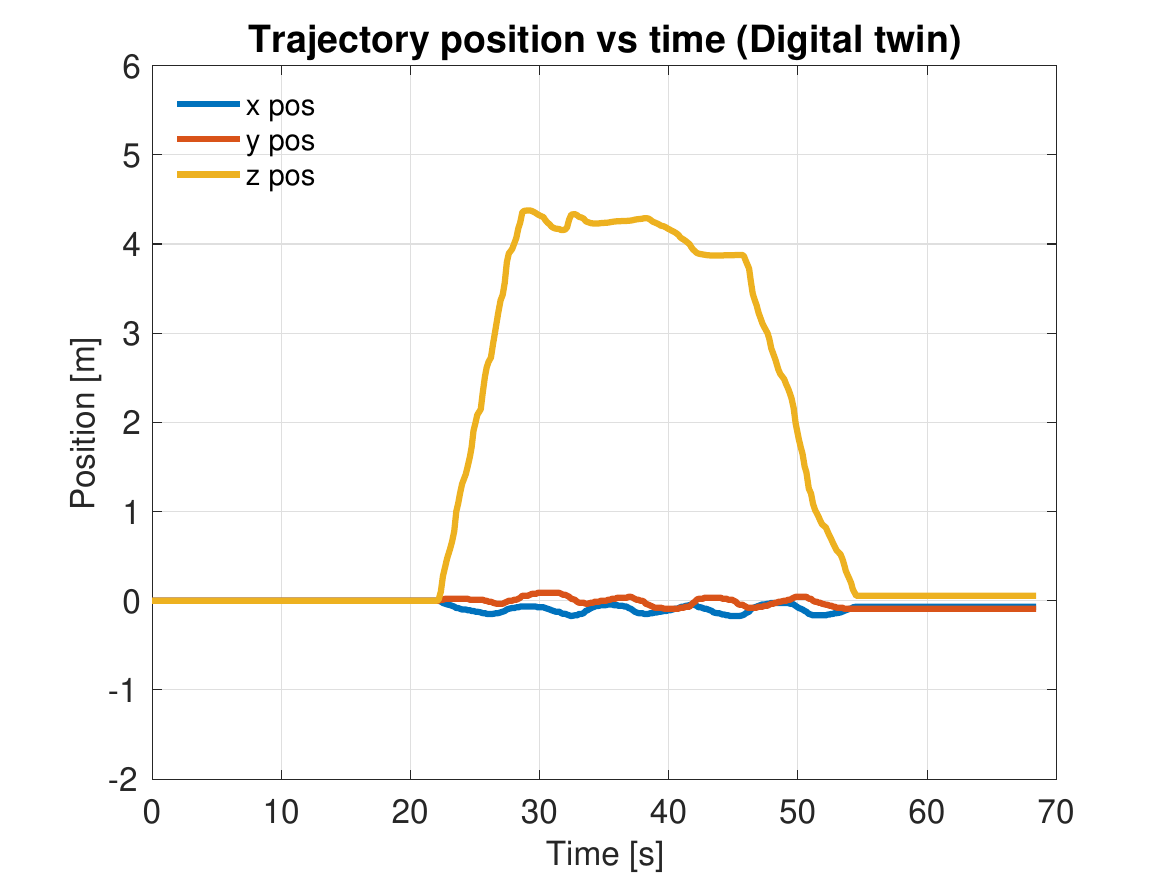}
        \caption{Position history of Task 1 using digital twin}
        \label{fig:task1posair}
    \end{subfigure}
    \hfill
    \begin{subfigure}[b]{0.48\columnwidth}
        \centering
        \includegraphics[width=\columnwidth]{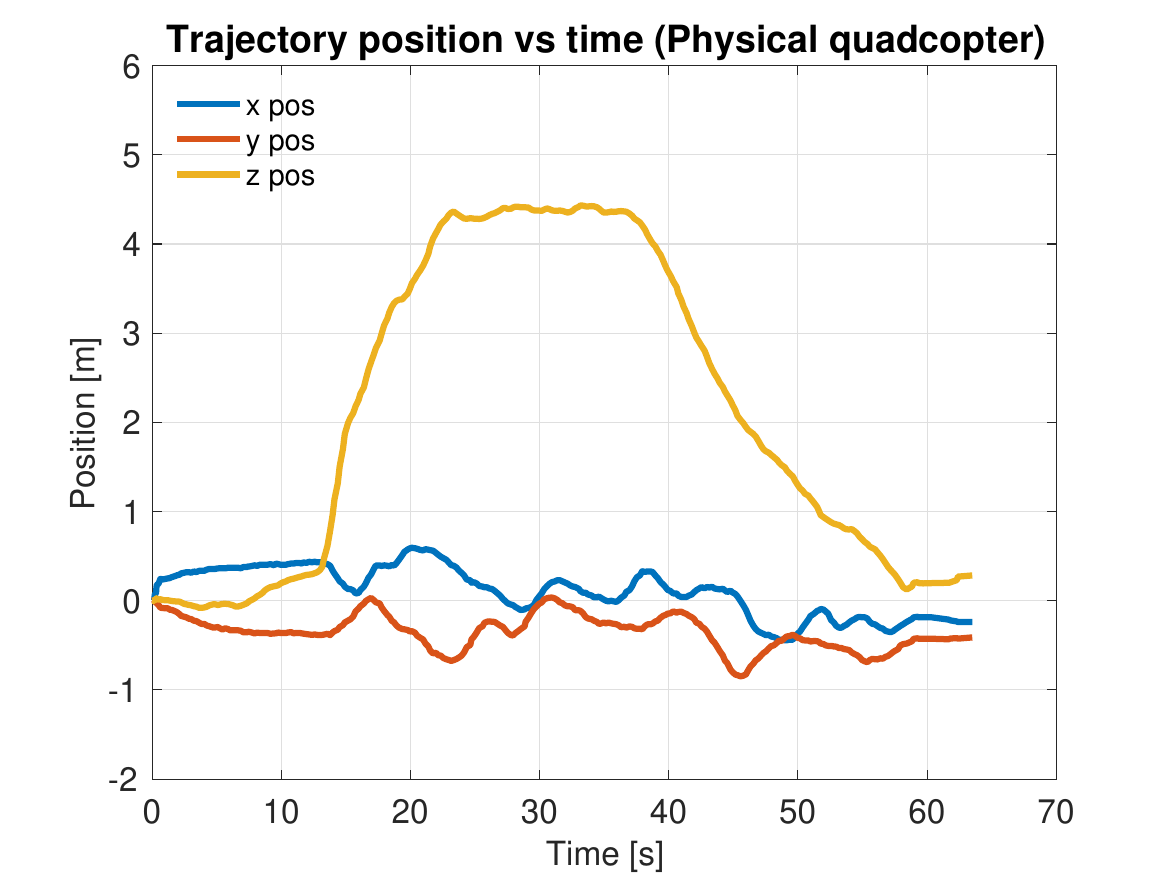}
        \caption{Position history of Task 1 using physical quadcopter}
        \label{fig:task1posout}
    \end{subfigure}
    \caption{Position history of Task 1 using both platforms}
    \label{fig:task1result2/2}
\end{figure}

\begin{figure} [H]
    \centering
    \begin{subfigure}[b]{0.48\columnwidth}
        \centering
        \includegraphics[width=\columnwidth]{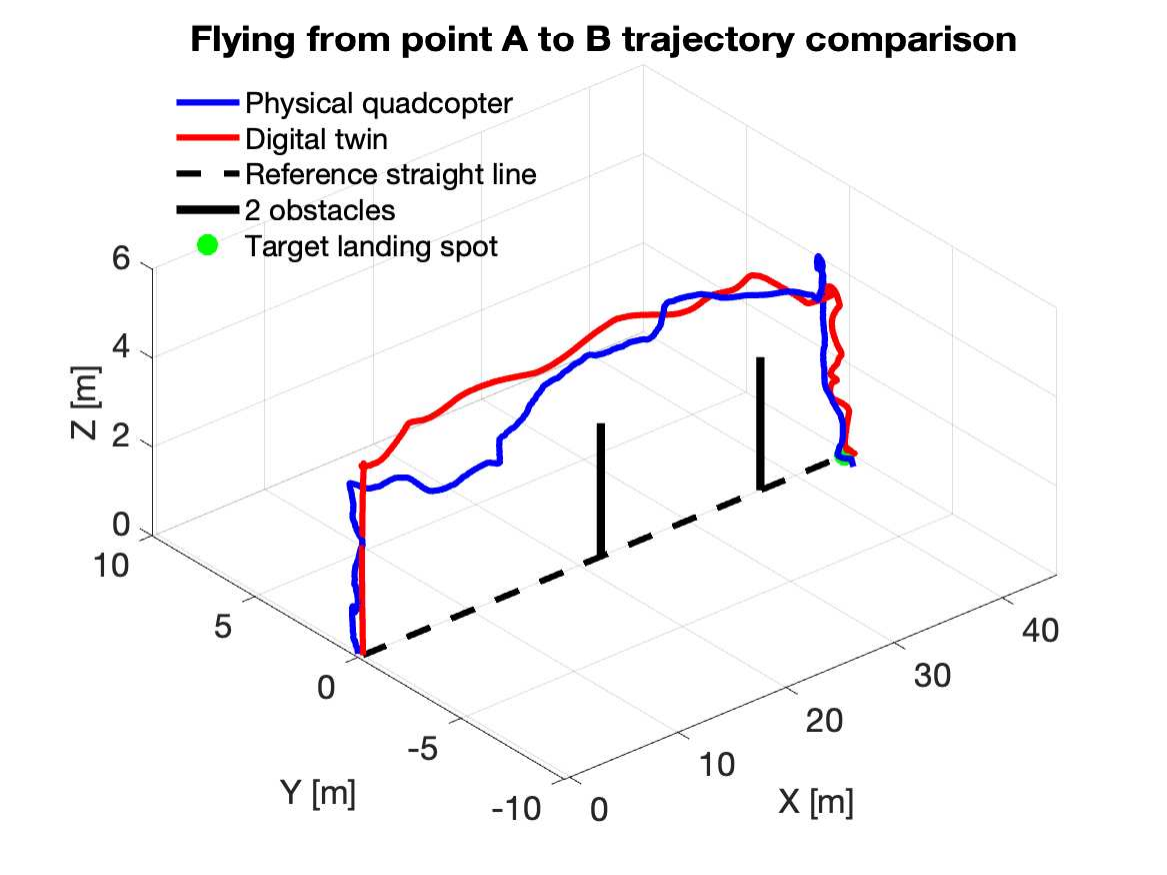}
        \caption{3D trajectory of Task 2 (both platforms)}
        \label{fig:3Dtask2}
    \end{subfigure}
    \hfill
    \begin{subfigure}[b]{0.48\columnwidth}
        \centering
        \includegraphics[width=\columnwidth]{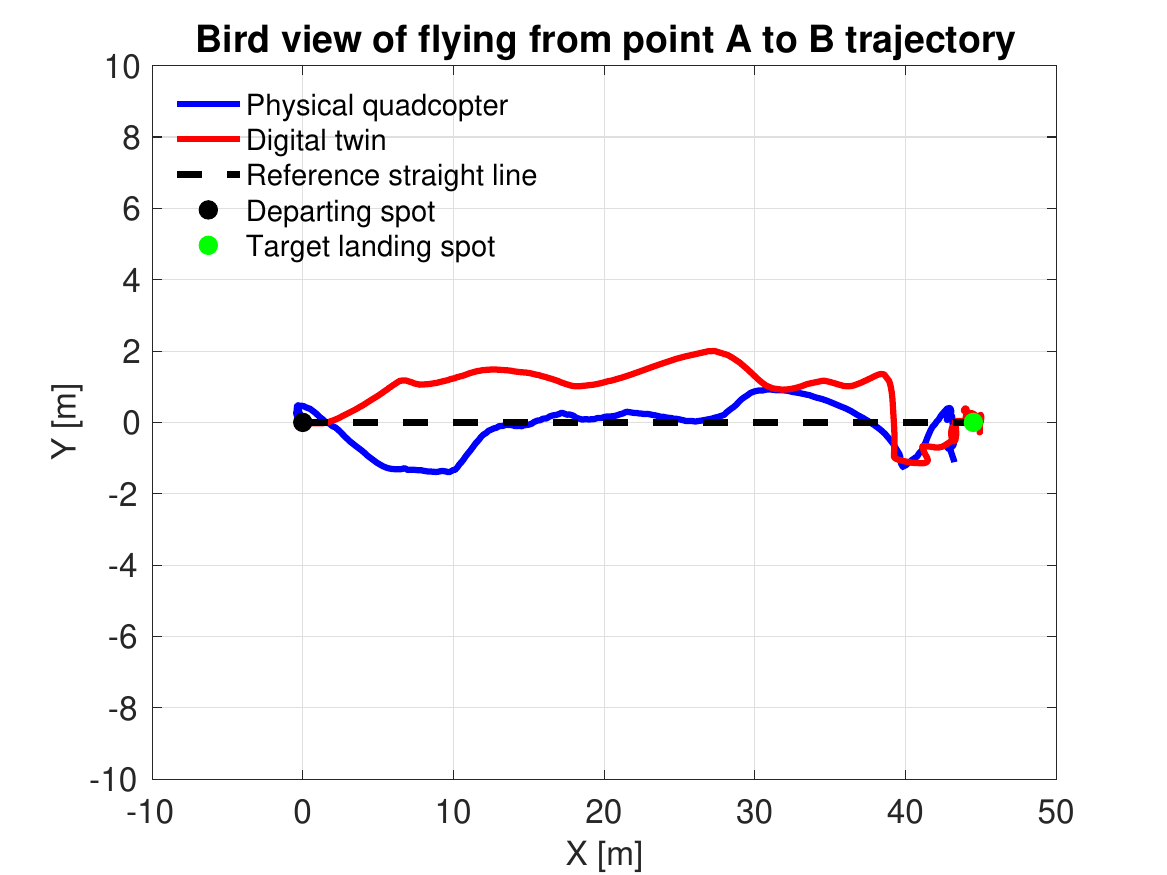}
        \caption{Bird-view trajectory of Task 2 (both platforms)}
        \label{fig:2Dtask2}
    \end{subfigure}
    \caption{Visualized flown trajectory of Task 2 using both platforms}
    \label{fig:task2result1/2}
\end{figure}

\begin{figure} [H]
    \centering
    \begin{subfigure}[b]{0.48\columnwidth}
        \centering
        \includegraphics[width=\columnwidth]{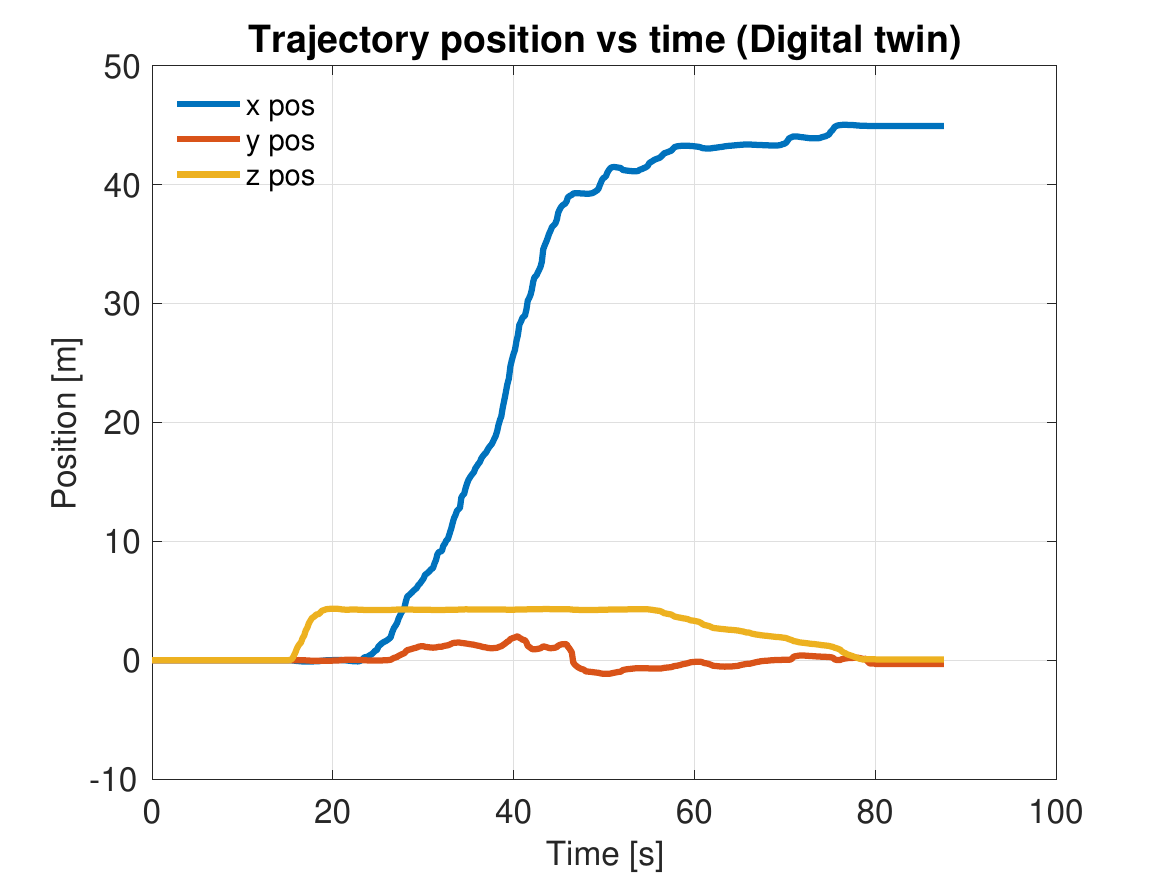}
        \caption{Position history of Task 2 using digital twin}
        \label{fig:task2posair}
    \end{subfigure}
    \hfill
    \begin{subfigure}[b]{0.48\columnwidth}
        \centering
        \includegraphics[width=\columnwidth]{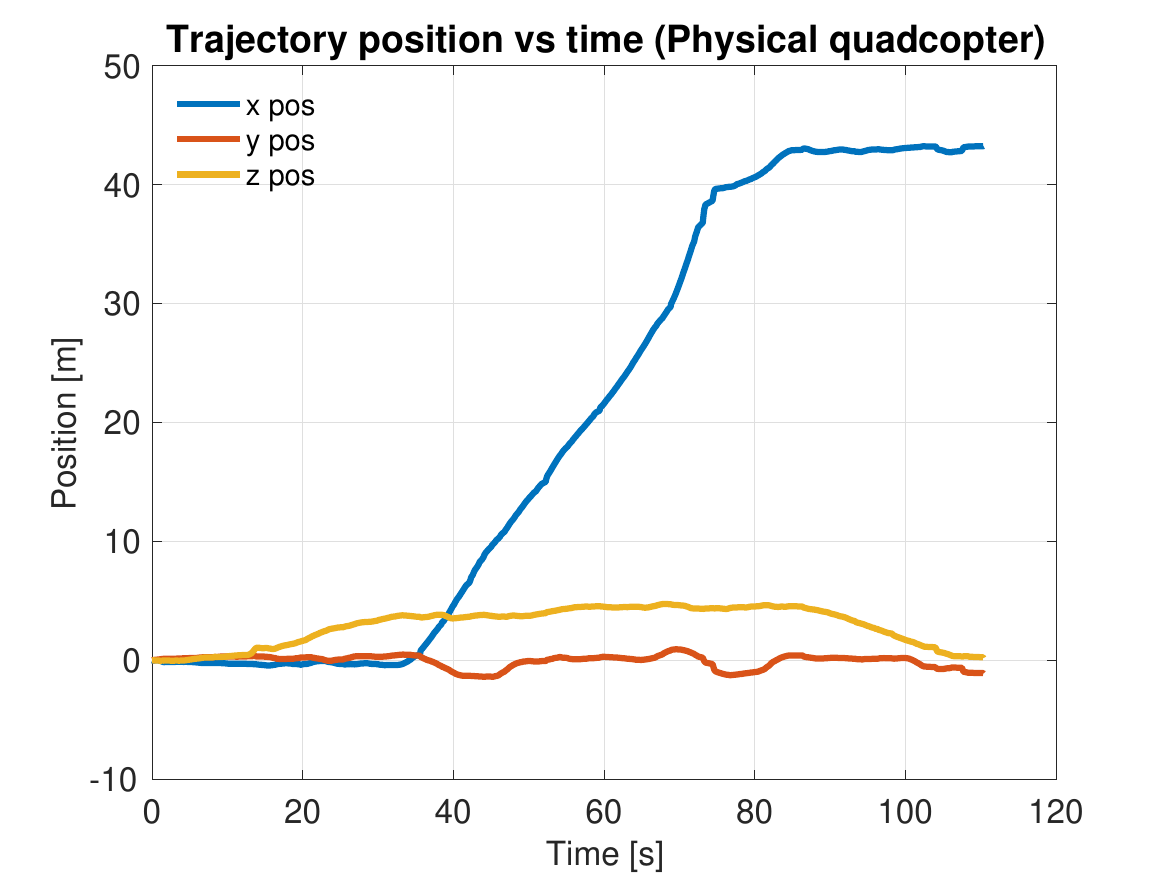}
        \caption{Position history of Task 2 using physical quadcopter}
        \label{fig:task2posout}
    \end{subfigure}
    \caption{Position history of Task 2 using both platforms}
    \label{fig:task2result2/2}
\end{figure}

\begin{figure} [H]
    \centering
    \begin{subfigure}[b]{0.48\columnwidth}
        \centering
        \includegraphics[width=\textwidth]{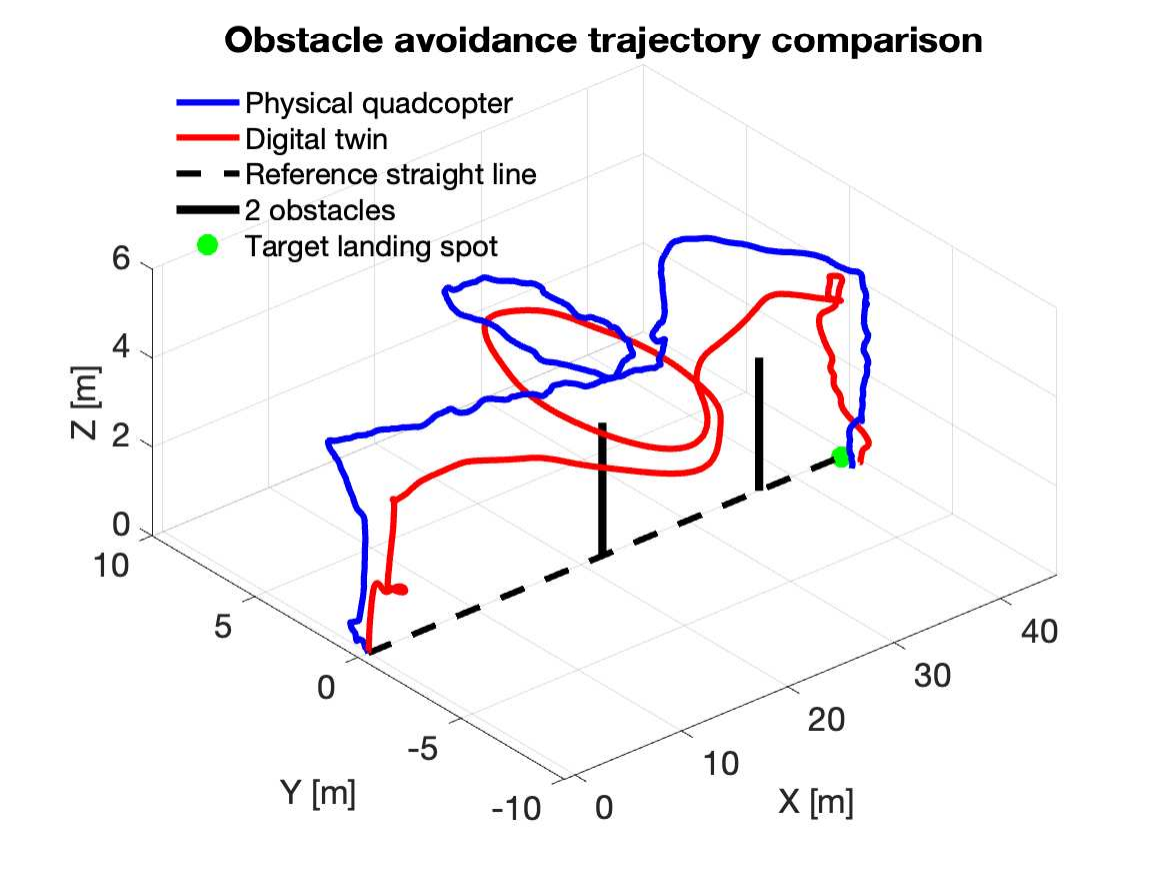}
        \caption{3D trajectory of Task 3 (both platforms)}
        \label{fig:3Dtask3}
    \end{subfigure}
    \hfill
    \begin{subfigure}[b]{0.48\columnwidth}
        \centering
        \includegraphics[width=\columnwidth]{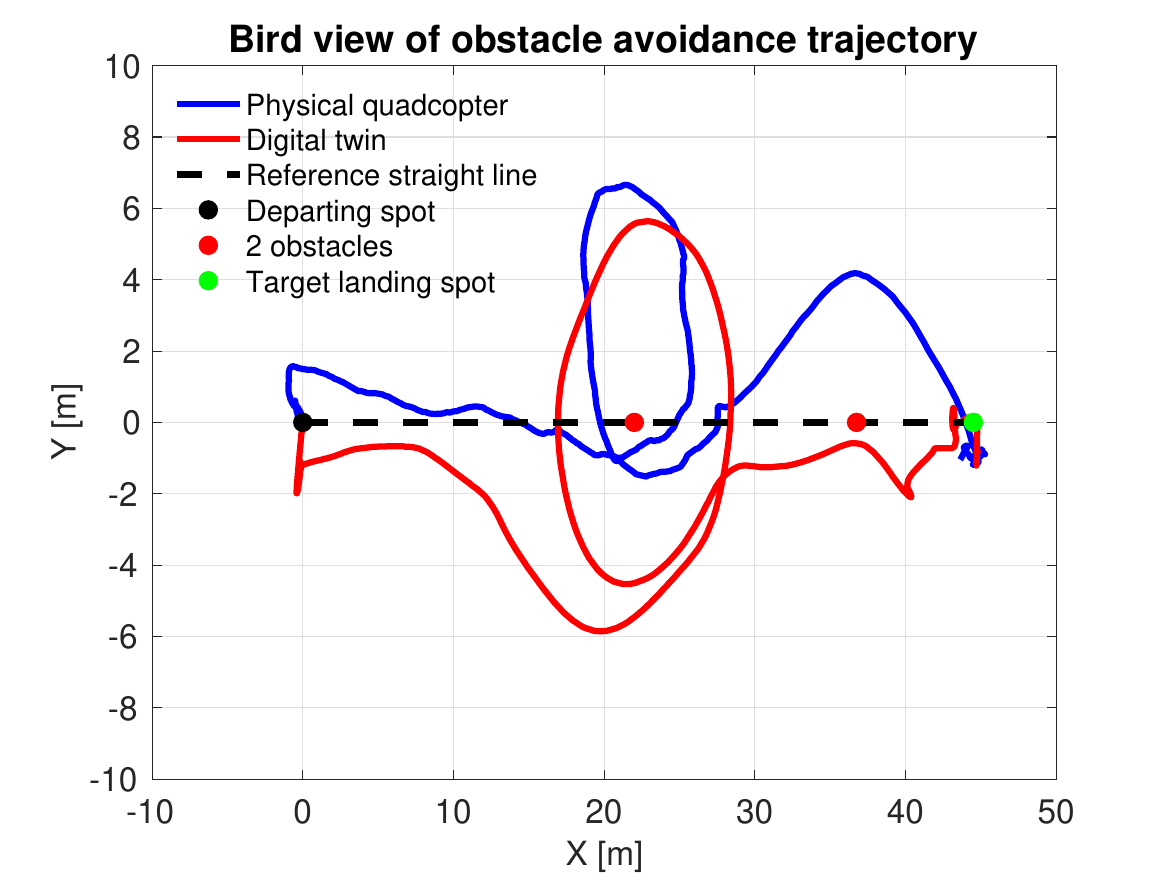}
        \caption{Bird-view trajectory of Task 3 (both platforms)}
        \label{fig:2Dtask3}
    \end{subfigure}
    \caption{Visualized flown trajectory of Task 3 using both platforms}
    \label{fig:task3result1/2}
\end{figure}

\begin{figure} [H]
    \centering
    \begin{subfigure}[b]{0.48\columnwidth}
        \centering
        \includegraphics[width=\columnwidth]{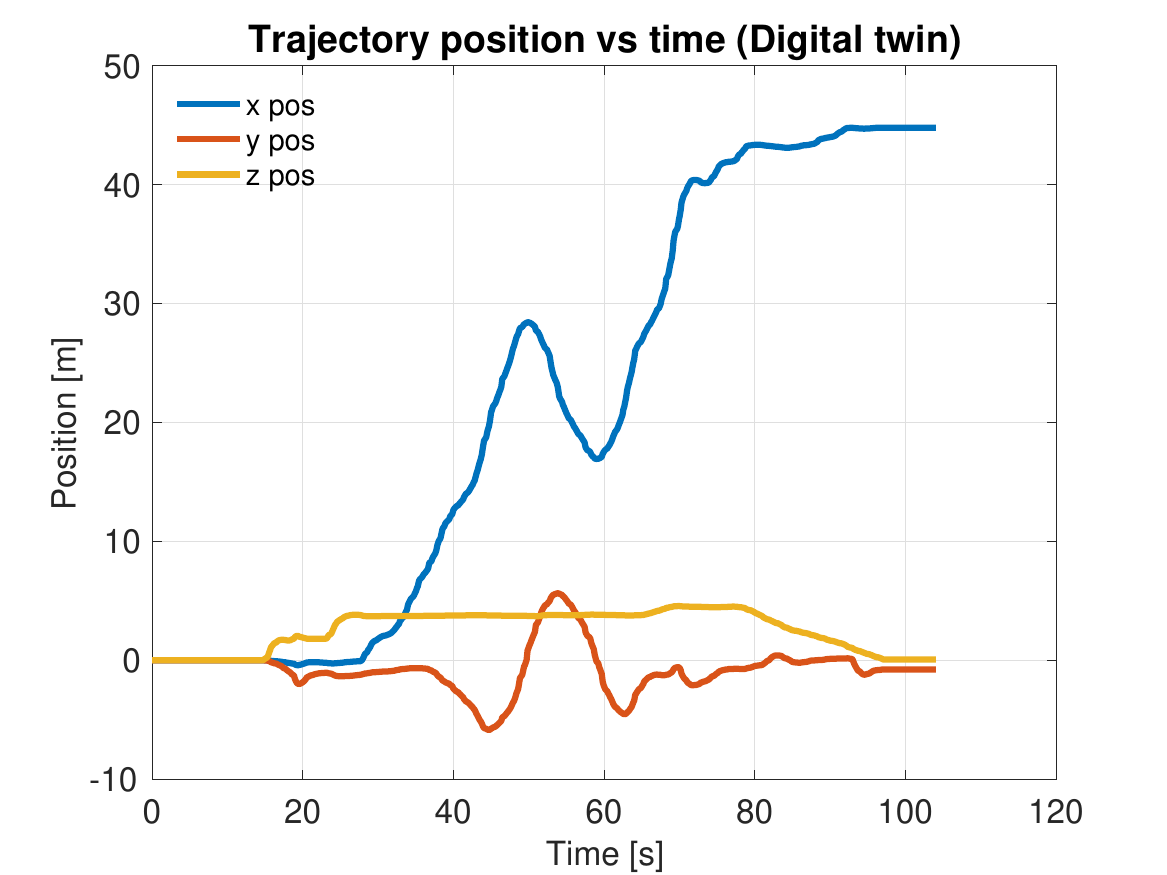}
        \caption{Position history of Task 3 using digital twin}
        \label{fig:task3posair}
    \end{subfigure}
    \hfill
    \begin{subfigure}[b]{0.48\columnwidth}
        \centering
        \includegraphics[width=\columnwidth]{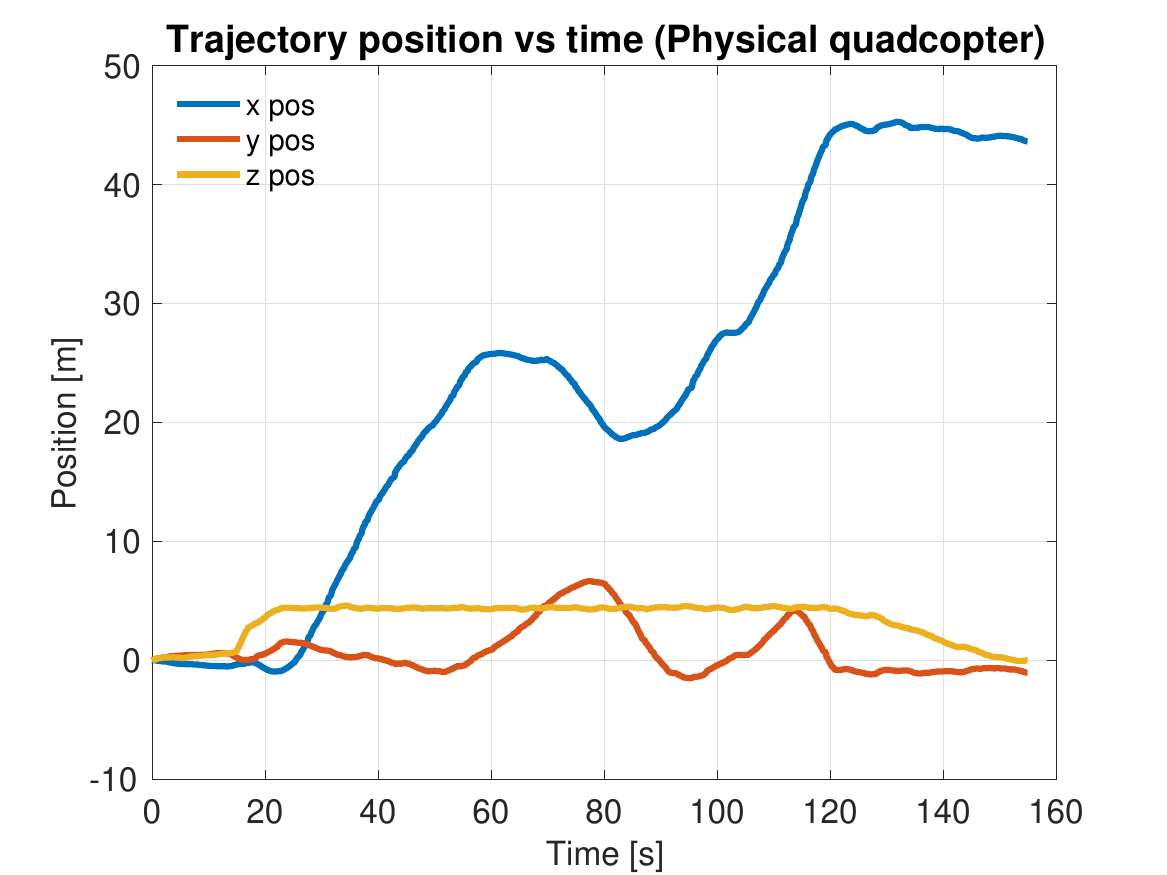}
        \caption{Position history of Task 3 using physical quadcopter}
        \label{fig:task3posout}
    \end{subfigure}
    \caption{Position history of Task 3 using both platforms}
    \label{fig:task2result3/2}
\end{figure}

\begin{figure} [H]
    \centering
    \begin{subfigure}[b]{0.48\columnwidth}
        \centering
        \includegraphics[width=\columnwidth]{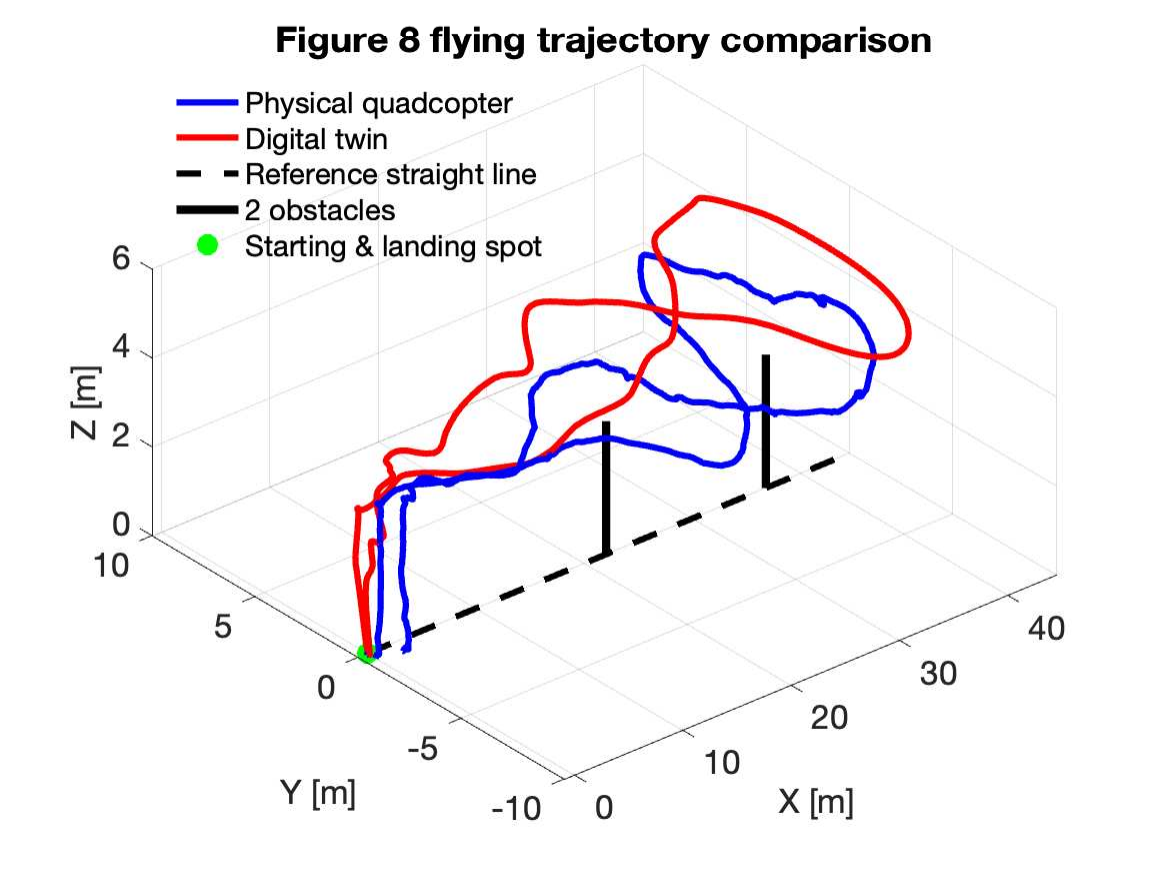}
        \caption{3D trajectory of Task 4 (both platforms)}
        \label{fig:3Dtask4}
    \end{subfigure}
    \hfill
    \begin{subfigure}[b]{0.48\columnwidth}
        \centering
        \includegraphics[width=\columnwidth]{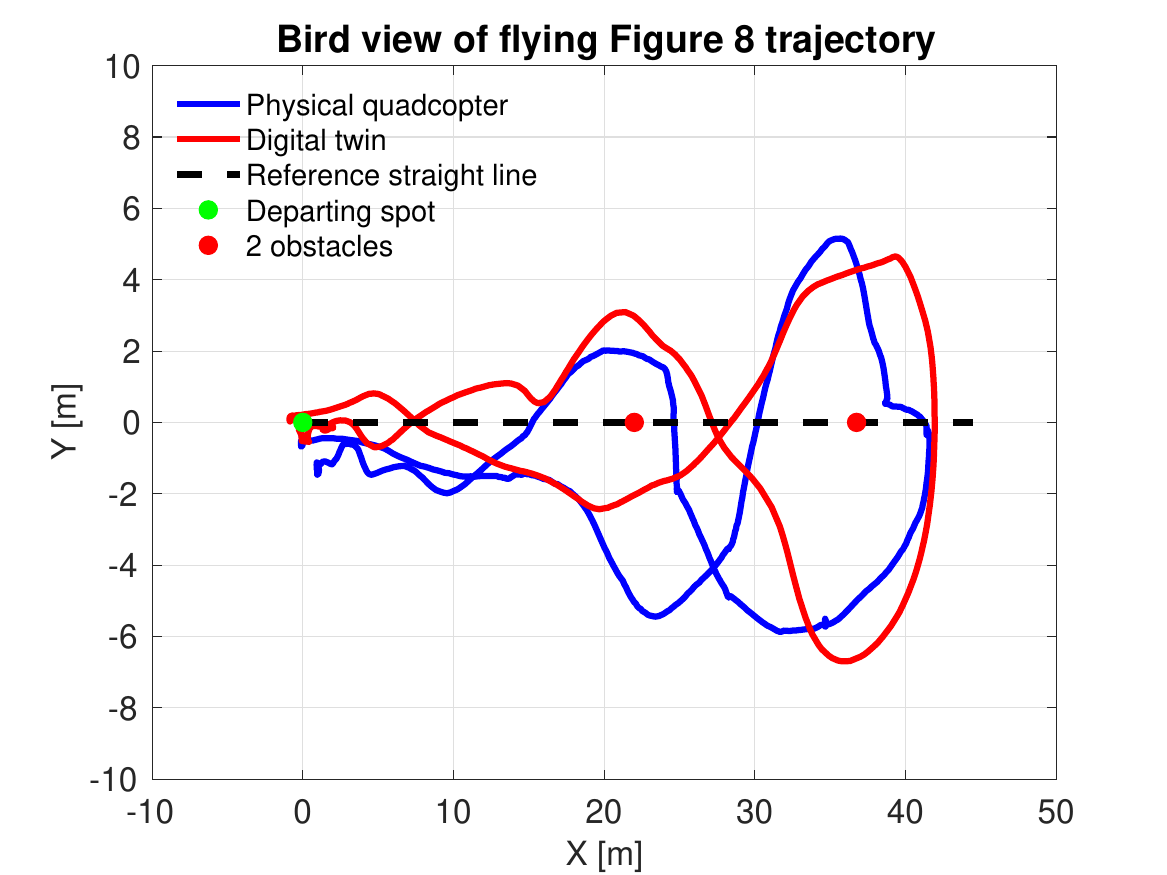}
        \caption{Bird-view trajectory of Task 4 (both platforms)}
        \label{fig:2Dtask4}
    \end{subfigure}
    \caption{Visualized flown trajectory of Task 4 using both platforms}
    \label{fig:task4result1/2}
\end{figure}

This section outlines the visualized results of the pilot's flying trajectories from the physical experiments using the physical quadcopter platform and the digital twin. By implementing the alignment method introduced in \ref{sec:GPScoordinate}, all GPS data points recorded along flight trajectories are mapped to align with due east, with the starting point as the initial position. For simplicity, this ENU frame is referred to as the XYZ frame. X is used to record longitudinal motion, Y is used to record lateral motion, and Z is used to record flight height variation during the flight. The visualized results, shown in Figure \ref{fig:task1result1/2}--\ref{fig:task4result3/2} then are demonstrated in both 3D view and bird-view, along with plots showing detailed position information corresponding to timestamps.

To analyze the pilot's maneuver behavior during the flights, the maneuvering portion of the data was extracted according to each of the flying tasks for statistical analysis. For task 1, which includes takeoff, hover, and landing, the maneuvering data was chosen between the point where the altitude first reached the target altitude of 4 meters and 10 seconds afterward. For the rest of the flying tasks associated with longitudinal motion, the maneuvering data was chosen between the longitudinal position of 2 meters after the task started and 2 meters before the task ended. The detailed statistic results are shown in table \ref{tab: task1}--\ref{tab: task4}. Please note that the platform used in outdoor experiments is the physical quadcopter, while the one used in the virtual environment is the digital twin.

\subsection{\bf Result visualization - Task 1}

\begin{table*}
\centering
\caption{Task 1 trails information using both platforms}
\label{tab: task1}
\begin{tabular}{ccccl}
\cline{1-4}
\multicolumn{4}{|c|}{Task 1: Take off, hover, and land flown trails information using both platforms}                                                                                          &  \\ \cline{1-4}
\multicolumn{1}{|c|}{Platform used}           & \multicolumn{1}{c|}{Hovering distance to origin (m)} & \multicolumn{1}{c|}{Height deviation (m)} & \multicolumn{1}{c|}{Trail time length (s)} &  \\ \cline{1-4}
\multicolumn{1}{|c|}{Physical quadcopter} & \multicolumn{1}{c|}{0.3691}                                  & \multicolumn{1}{c|}{0.0733}               & \multicolumn{1}{c|}{63.4972}        &  \\ \cline{1-4}
\multicolumn{1}{|c|}{Digital twin}        & \multicolumn{1}{c|}{0.1056}                                  & \multicolumn{1}{c|}{0.0693}               & \multicolumn{1}{c|}{68.4444}        &  \\ \cline{1-4}
\multicolumn{1}{l}{}                          & \multicolumn{1}{l}{}                                         & \multicolumn{1}{l}{}                      & \multicolumn{1}{l}{}                & 
\end{tabular}
\end{table*}

\subsection{\bf Result visualization - Task 2}

\begin{table*}
\centering
\caption{Task 2 trails information using both platforms}
\label{tab: task2}
\begin{tabular}{ccccl}
\cline{1-4}
\multicolumn{4}{|c|}{Task 2: Flying from point A to B flown trails information using both platforms}                                                                          &  \\ \cline{1-4}
\multicolumn{1}{|c|}{Platform used}           & \multicolumn{1}{c|}{Lateral deviation (m)} & \multicolumn{1}{c|}{Height deviation (m)} & \multicolumn{1}{c|}{Trail time length (s)} &  \\ \cline{1-4}
\multicolumn{1}{|c|}{Physical quadcopter} & \multicolumn{1}{c|}{0.6728}                & \multicolumn{1}{c|}{0.3654}               & \multicolumn{1}{c|}{110.3442}       &  \\ \cline{1-4}
\multicolumn{1}{|c|}{Digital twin}        & \multicolumn{1}{c|}{1.0038}                & \multicolumn{1}{c|}{0.0302}               & \multicolumn{1}{c|}{87.5617}        &  \\ \cline{1-4}
\multicolumn{1}{l}{}                          & \multicolumn{1}{l}{}                       & \multicolumn{1}{l}{}                      & \multicolumn{1}{l}{}                & 
\end{tabular}
\end{table*}

\subsection{\bf Result visualization - Task 3}

\begin{table*}
\centering
\caption{Task 3 trails information using both platforms}
\label{tab: task4}
\begin{tabular}{ccccl}
\cline{1-4}
\multicolumn{4}{|c|}{Task 3: Obstacle avoidance flown trails information using both platforms}                                                                                &  \\ \cline{1-4}
\multicolumn{1}{|c|}{Platform used}           & \multicolumn{1}{c|}{Lateral deviation (m)} & \multicolumn{1}{c|}{Height deviation (m)} & \multicolumn{1}{c|}{Trail time length (s)} &  \\ \cline{1-4}
\multicolumn{1}{|c|}{Physical quadcopter} & \multicolumn{1}{c|}{2.3556}                & \multicolumn{1}{c|}{0.0648}               & \multicolumn{1}{c|}{154.9210}       &  \\ \cline{1-4}
\multicolumn{1}{|c|}{Digital twin}        & \multicolumn{1}{c|}{2.7866}                & \multicolumn{1}{c|}{0.3042}               & \multicolumn{1}{c|}{104.0227}       &  \\ \cline{1-4}
\multicolumn{1}{l}{}                          & \multicolumn{1}{l}{}                       & \multicolumn{1}{l}{}                      & \multicolumn{1}{l}{}                & 
\end{tabular}
\end{table*}

\subsection{\bf Result visualization - Task 4}
\begin{figure} [H]
    \centering
    \begin{subfigure}[b]{0.48\columnwidth}
        \centering
        \includegraphics[width=\columnwidth]{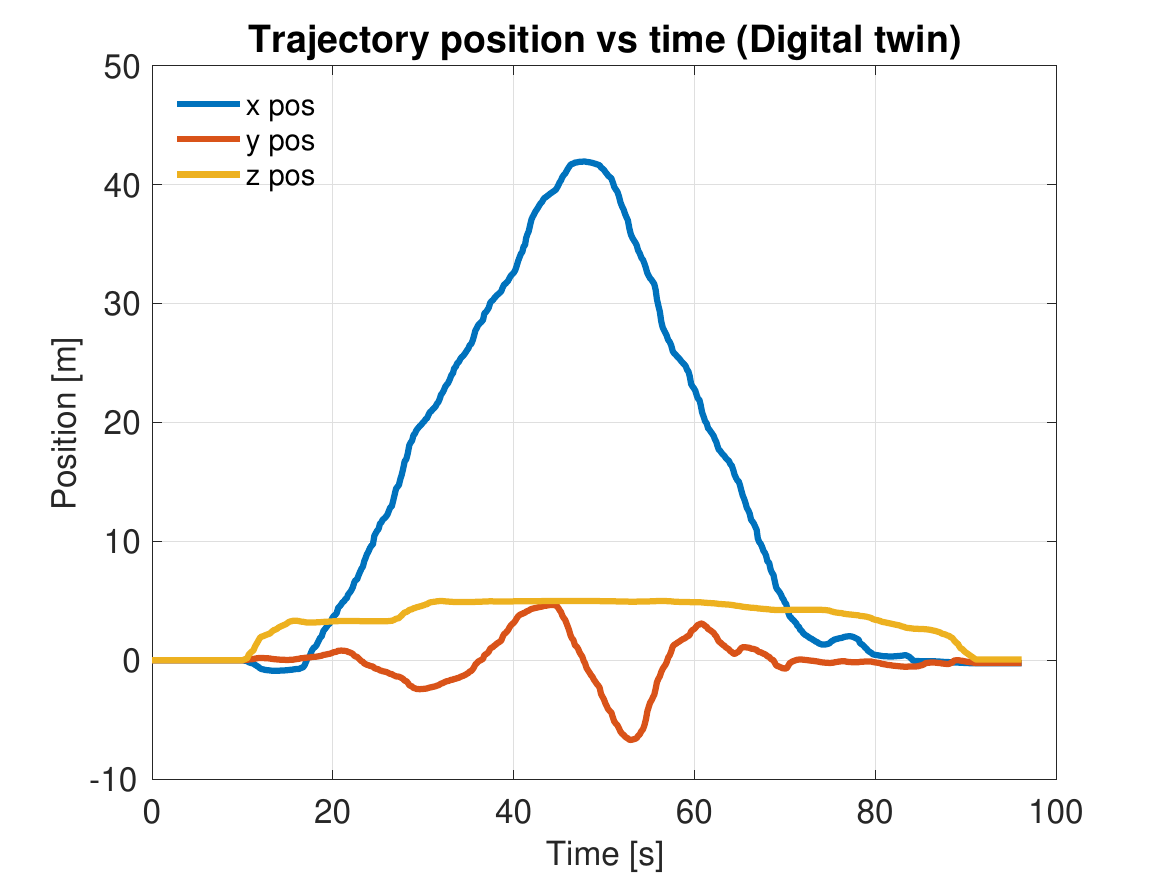}
        \caption{Position history of Task 4 in digital twin}
        \label{fig:task4posair}
    \end{subfigure}
    \hfill
    \begin{subfigure}[b]{0.48\columnwidth}
        \centering
        \includegraphics[width=\columnwidth]{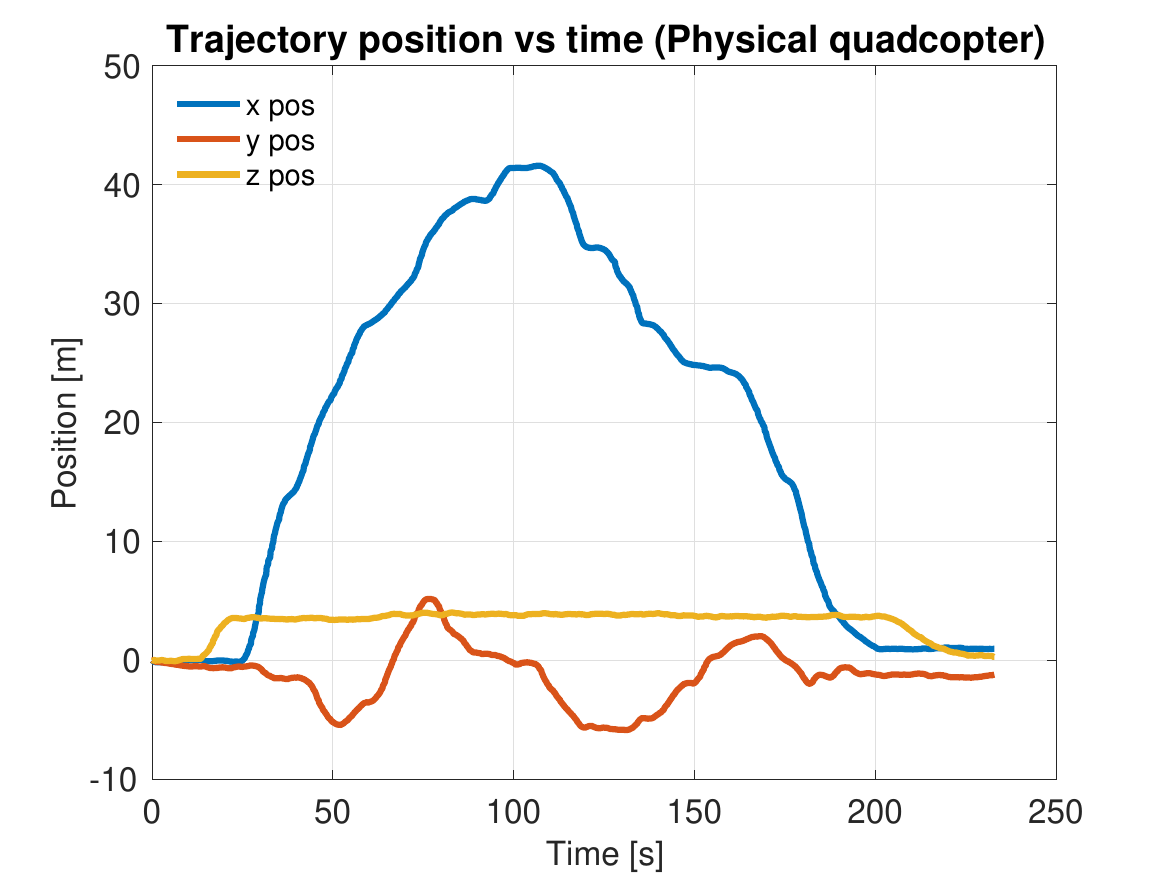}
        \caption{Position history of Task 4 in physical quadcopter}
        \label{fig:task4posout}
    \end{subfigure}
    \caption{Position history of Task 4 using both platforms}
    \label{fig:task4result3/2}
\end{figure}

\begin{table*}
\caption{Task 4 trails information using both platforms}
\label{tab: task4}
\footnotesize
\centering
\begin{tabular}{ccccl}
\cline{1-4}
\multicolumn{4}{|c|}{Task 4: Flying Figure 8 flown trails information using both platforms}                                                                                   &  \\ \cline{1-4}
\multicolumn{1}{|c|}{Platform used}           & \multicolumn{1}{c|}{Lateral deviation (m)} & \multicolumn{1}{c|}{Height deviation (m)} & \multicolumn{1}{c|}{Trail time length (s)} &  \\ \cline{1-4}
\multicolumn{1}{|c|}{Physical quadcopter} & \multicolumn{1}{c|}{2.7589}                & \multicolumn{1}{c|}{0.1579}               & \multicolumn{1}{c|}{232.8868}       &  \\ \cline{1-4}
\multicolumn{1}{|c|}{Digital twin}        & \multicolumn{1}{c|}{2.4729}                & \multicolumn{1}{c|}{0.5843}               & \multicolumn{1}{c|}{96.1778}        &  \\ \cline{1-4}
\multicolumn{1}{l}{}                          & \multicolumn{1}{l}{}                       & \multicolumn{1}{l}{}                      & \multicolumn{1}{l}{}                & 
\end{tabular}
\label{}
\end{table*}
\subsection{\bf Discussions} 
The statistical results demonstrated in Tables \ref{tab: task1}--\ref{tab: task4} indicate the maneuver behavior differences between the same pilot using the physical platform and the digital twin in all four flying tasks. Task 1 results shown in Table \ref{tab: task1} indicate that the average distance between the hovering positions and origin using the digital twin is 0.1056 meters; while using the physical platform, it is 0.3691 meters during the hovering period. The hovering position difference reflects how far the pilot drifted using both platforms and flying using the digital twin demonstrated slightly better hovering control. Unlike Task 1, Tasks 2 to 4 require the pilot to manage both longitudinal and lateral control with increasing complexity. Tables \ref{tab: task2}--\ref{tab: task4} demonstrate the standard deviation of the trajectory Y and Z positions when maneuvering, reflecting the variation in lateral position and height during the flight, along with the time taken to complete the task trails.
As the flying task complexity increases, the time required for the pilot using the physical quadcopter to finish the trail shows an increasing trend from 110.3442 seconds to 232.8868 seconds. Additionally, the lateral deviation increases from 0.6728 meters to 2.7589 meters. These differences indicate that the pilot made larger turns in Task 4 than in Task 2 and Task 3 while maneuvering with the physical platform. Conversely, the pilot’s Task 4 maneuver performance using the digital twin in the Airsim simulation demonstrated less time consumption with a smaller lateral deviation than his result in Task 3. This suggests that when flying in AirSim simulation, as lateral motion requirements increase, the digital twin controlled through Pixhawk is easier to manage than the real physical quadcopter.

Even though both the physical quadcopter platform and the digital twin embedded in the Airsim simulation share the same controller firmware, Pixhawk PX4, and both platforms are flown by the same pilot in identical flying tasks, discrepancies are still observed in the analysis of the results. Considering that Pixhawk 2.4.8 is no longer maintained for firmware updates, sensor messages for odometry information cannot be extracted through Pymavlink to visualize the trajectory directly. Therefore, the trajectory visualization is based solely on GPS measurements taken during the flights. GPS noise in outdoor experiments could potentially explain why task 1 (takeoff, hover, and land) shows slightly poorer motion control performance using the physical quadcopter compared to the digital twin in the Airsim simulation.

Additionally, the difference in software architecture between the GUI for the physical quadcopter framework and the virtual GUI for the digital twin in Airsim simulation may have contributed to the performance discrepancy in tasks 3 and 4. The GUI for the physical quadcopter embeds the local network communication in the real-time data extraction process. Even though reading Pixhawk data through a serial connection can make the data transmission frequency much faster than telemetry, extra encoding and decoding processes, along with the local network’s transmission and receiving speed, can add delay. Through testing, considering the factor of local network connection, an extra delay between 0.015 and 0.3 second was observed at every data refresh iteration, from the moment that real-time Pixhawk data was read by the socket server until it was received and decoded by the local GUI client and ready to be queued for update. However, in the Airsim simulation, the virtual GUI interface is free from the extra process of socket communication. This means that the flying information in the digital twin’s FPV interface is in absolute real-time and free from the delay caused by data encoding/decoding or the limitations of local network transmission speed. Consequently, the pilot needs more time to make decisions in more complex flying tasks using the physical quadcopter than when using the digital twin. This can be reflected in the time length of the X curve being flat in Figure \ref{fig:task4posout}.

\subsection{\bf Data Processing for Reinforcement Learning}
Considering the physical quadcopter and the digital twin detailed in this paper, there exists ample scope to integrate these tools with RL Libraries for the development of autonomous UAVs. Flight data from the physical quadcopter provides valuable insights into real-world flight dynamics and environmental interactions, while the digital twin enables risk-free data collection and testing across diverse scenarios.
The collected data includes telemetry, GPS, control inputs, and dual-view camera frames captured during flights using the physical quadcopter and digital twin. These datasets capture the complex maneuvers of expert pilots. Before training an autonomous UAV using this data, pre-processing is crucial. The input stream must be synchronized to ensure that each training instance accurately represents the UAV's state and corresponding pilot actions at a specific moment in time. This involves aligning timestamps across sensor readings, control inputs, and camera images. Furthermore, normalizing state variables ensures consistent input ranges for the learning algorithms.
\subsubsection{Custom Gym Environment}
To effectively utilize RL algorithms, an environment should be developed following the standard framework established by Gymnasium \cite{towers_gymnasium_2023}. Developing this allows easy integration with any popular open-source RL libraries such as  
stable-baselines3 \cite{stable-baselines3} or RLlib \cite{liang2018rllib}. This Gym environment leverages defined state and action spaces that reflect real-world operational parameters with the reward formulation for each step.  An example custom environment is developed and is also made available in our GitHub repository. The state and action space for the environment is as follows, 
\begin{itemize}
    \item \textbf{State Space}
    Includes spatial coordinates (latitude, longitude, altitude), velocities, and orientations derived from GPS and IMU sensors. Visual inputs from FPV camera and bottom camera. Additionally, navigational data, such as relative distances to target locations, are also integrated.
    \item \textbf{Action Space}
    Actions include control inputs such as throttle, pitch, yaw, and roll adjustments, which directly affect the quadcopter's moments.
\end{itemize}
Note that the above state and action space is an example to show the API level connections to gather the collected experiment data and this should be modified based on the requirements. 

\section{\bf CONCLUSIONS}\label{sec:conclusion}

In this paper, we introduced a low-cost hardware/software architecture developed based on the F450 quadcopter frame to support human FPV flight demonstrations with comprehensive low-level data logging capabilities. The underlying software architecture, developed using the Python-based Kivy library, ensures that real-time data read from the Pixhawk flight controller and image frames from dual-view FPV cameras can be concurrently displayed with synchronized data logging by leveraging socket communication and multiprocessing through Kivy’s unique non-blocking clock event updates. In addition, we employed a virtual FPV quadcopter within a scaled testing ground, as a digital twin for flights conducted in the SOAR outdoor space at the university. This is developed using AirSim/Unreal Engine-based simulations along with a HITL setup via Pixhawk PX4 firmware, which is identical to one implemented on the hardware architecture. With the two FPV quadcopter platforms, we conducted four flying experiments with varying levels of complexity. Analyses of the flights experiment data showed that the pilot demonstrated better hovering and lateral control with frequent turning requirements and longitudinal motion when using the digital twin based on the AirSim HITL simulation. In contrast, in the physical flight experiments, better altitude maintenance was demonstrated as the flying task complexity increased.

This platform can be useful to rigorously test and validate new AI-based algorithms in a safe and realistic simulation environment before deploying them on physical quadcopters or larger aircraft, which can reduce the risk of crashing in real-world testing and shorten the development time. We have also proposed a custom environment as an example for implementing collected data in RL algorithms. To improve the quality of data fed to the RL environment, future work could involve exploring methods to enhance the frequency of data acquisition and adding additional modules to verify the accuracy of the saved GPS coordinates. Thus, the demonstrated capabilities and initial flight test findings thereof points to both the need for and the feasibility of building open-source hardware/software platforms and digital twins to reduce barriers to entry for (and push forward) research in aerial autonomy.

\addtolength{\textheight}{-12cm}   





\bibliographystyle{IEEEtran}
\bibliography{pr}

\end{document}